\documentclass[acmlarge]{acmart}

\usepackage{multirow} 
\usepackage{array}

\usepackage{amsmath}
\usepackage{amssymb}
\usepackage{mathtools}
\usepackage{amsthm}

\usepackage{subfiles} 
\usepackage{makecell} 
\usepackage{multirow} 
\usepackage{subcaption}
\usepackage{color,xcolor}
\usepackage{graphicx}
\usepackage{svg}
\usepackage{booktabs}
\usepackage{makecell}
\usepackage{colortbl}
\usepackage{pifont}
\usepackage{algorithm} 
\usepackage{listings}
\usepackage{algorithmic} 
\usepackage[switch]{lineno}
\usepackage{amssymb} 

\usepackage[misc]{ifsym}

\usepackage[normalem]{ulem}

\usepackage[capitalize,noabbrev]{cleveref}
\usepackage{tcolorbox}
\usepackage{hyperref}
\usepackage{longtable}
\usepackage{arydshln}

\theoremstyle{plain}

\theoremstyle{definition}

\theoremstyle{remark}

\usepackage[textsize=tiny]{todonotes}

\definecolor{shapecolor}{rgb}{0.0,0.5,0.0}

\definecolor{arylideyellow}{rgb}{0.91, 0.84, 0.42}

\AtBeginDocument{%
  }

\setcopyright{acmlicensed}
\copyrightyear{2018}
\acmJournal{IMWUT}
\acmYear{2025}
\acmMonth{9}
\acmVolume{0}
\acmNumber{0}
\acmDOI{XXXXXXX.XXXXXXX}
\acmISBN{978-1-4503-XXXX-X/18/06}

\begin{document}

\title{XRF V2: A Dataset for Action Summarization with Wi-Fi Signals, and IMUs in Phones,
Watches, Earbuds, and Glasses}

\renewcommand{\shortauthors}{XRF V2}


\author{Bo Lan}
\orcid{0009-0004-5777-6181}
\affiliation{
 \institution{School of Software Engineering, Xi'an Jiaotong University}
 \city{Xi'an}
 \state{Shaanxi}
 \country{China}
 }
  \affiliation{
 \institution{State Key Laboratory of Integrated Services Networks, 
Xidian University}
 \city{Xi'an}
 \state{Shaanxi}
 \country{China}}
 \email{bolan@stu.xjtu.edu.cn}

\author{Pei Li}
\orcid{0009-0006-3194-556X}
\affiliation{
  \institution{School of Software Engineering, Xi'an Jiaotong University}
 \city{Xi'an}
 \state{Shaanxi}
 \country{China}}
 \email{lp@stu.xjtu.edu.cn}

\author{Jiaxi Yin}
\orcid{0009-0009-5632-7890}
\affiliation{
 \institution{School of Software Engineering, Xi'an Jiaotong University}
 \city{Xi'an}
 \state{Shaanxi}
 \country{China}}
\email{jiaxiyin@stu.xjtu.edu.cn}

\author{Yunpeng Song}
\orcid{0000-0002-4186-0408}
\affiliation{
 \institution{MOE KLINNS Lab, Xi'an Jiaotong University}
 \city{Xi'an}
 \state{Shaanxi}
 \country{China}}
\email{yunpengs@xjtu.edu.cn}

\author{Ge Wang}
\orcid{0000-0002-9058-8543}
\affiliation{
 \institution{School of Computer Science and Technology, Xi'an Jiaotong University}
 \city{Xi'an}
 \state{Shaanxi}
 \country{China}}
\email{gewang@xjtu.edu.cn}

\author{Han Ding}
\orcid{0000-0002-5274-7988}
\affiliation{
 \institution{School of Computer Science and Technology, Xi'an Jiaotong University}
 \city{Xi'an}
 \state{Shaanxi}
 \country{China}}
\email{dinghan@xjtu.edu.cn}

\author{Jinsong Han}
\orcid{0000-0001-5064-1955}
\affiliation{
 \institution{College of Computer Science and Technology, Zhejiang University}
 \city{Hangzhou}
 \state{Zhejiang}
 \country{China}}
\email{hanjinsong@zju.edu.cn}

\author{Fei Wang}
\thanks{$*$ Fei Wang is the corresponding author who initiated and led the project. Bo Lan and Pei Li contributed equally. }
\orcid{0000-0002-0750-6990}
\authornotemark[1]
\affiliation{
 \institution{School of Software Engineering, Xi'an Jiaotong University}
 \city{Xi'an}
 \state{Shaanxi}
 \country{China}
 }
 \affiliation{
 \institution{State Key Laboratory of Integrated Services Networks, 
Xidian University}
 \city{Xi'an}
 \state{Shaanxi}
 \country{China}}
\email{feynmanw@xjtu.edu.cn}

\begin{abstract}
Human Action Recognition (HAR) plays a crucial role in applications such as health monitoring, smart home automation, and human-computer interaction. While HAR has been extensively studied, action summarization using Wi-Fi and IMU signals in smart-home environments , which involves identifying and summarizing continuous actions, remains an emerging task. This paper introduces the novel XRF V2 dataset, designed for indoor daily activity Temporal Action Localization (TAL) and action summarization. XRF V2 integrates multimodal data from Wi-Fi signals, IMU sensors (smartphones, smartwatches, headphones, and smart glasses), and synchronized video recordings, offering a diverse collection of indoor activities from 16 volunteers across three distinct environments. To tackle TAL and action summarization, we propose the XRFMamba neural network, which excels at capturing long-term dependencies in untrimmed sensory sequences and achieves the best performance with an average mAP of 78.74, outperforming the recent WiFiTAD by 5.49 points in mAP@avg while using 35\% fewer parameters. In action summarization, we introduce a new metric, Response Meaning Consistency (RMC), to evaluate action summarization performance. And it achieves an average Response Meaning Consistency (mRMC) of 0.802. We envision XRF V2 as a valuable resource for advancing research in human action localization, action forecasting, pose estimation, multimodal foundation models pre-training, synthetic data generation, and more. The data and code are available at~\href{https://github.com/aiotgroup/XRFV2}{\textcolor[RGB]{255, 69, 69}{https://github.com/aiotgroup/XRFV2}}.
\end{abstract}



\received{20 February 2007}
\received[revised]{12 March 2009}
\received[accepted]{5 June 2009}


\maketitle

\section{Introduction}\label{sec:introduction}

\begin{figure}[ht]
  \includegraphics[width=0.98\textwidth]{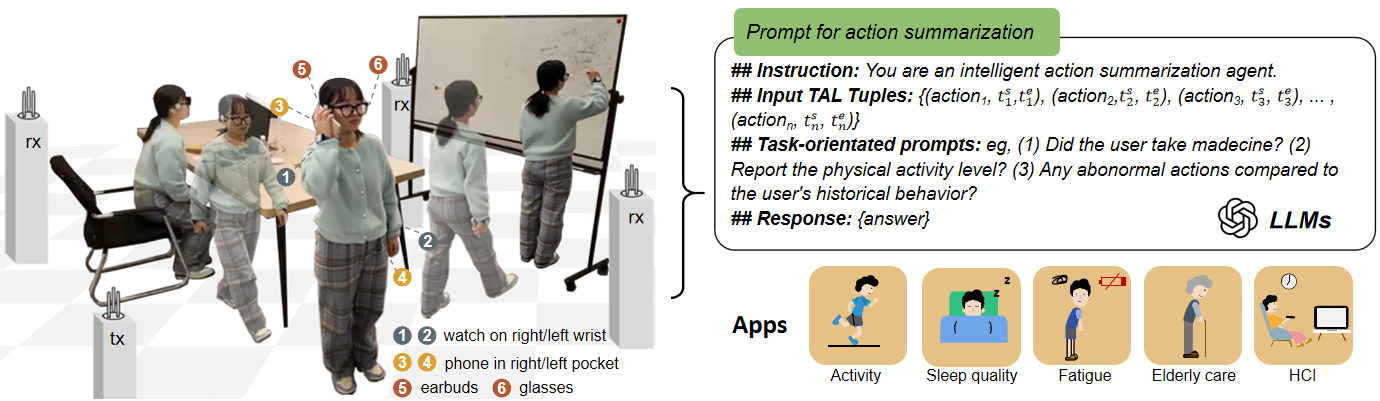}
  \caption{XRF V2 includes multimodal data from Wi-Fi signals, IMU sensors (smartphones, smartwatches, headphones, and smart glasses), and synchronized video recordings. Volunteers perform continuous actions indoors, and Temporal Action Localization (TAL) techniques are used to identify the start and end times of each action within untrimmed sequences, which are then combined into tuples. These action sequence tuples, in conjunction with task-oriented prompts, are input into LLMs, thereby enabling the LLMs to function as intelligent agents, supporting various applications, such as remote healthcare monitoring, personal assistant, smart home automation, and health and wellness analysis. }
  \label{fig:fig1}
\end{figure}

Human Indoor Action Understanding (HIAU) is pivotal in enabling intelligent environments, with wide-ranging applications in health monitoring, smart home automation, and human-computer interaction. Currently, human action recognition (HAR) is the most extensively studied aspect in HIAU. HAR aims to identify and categorize discrete human actions by leveraging data from diverse modal sensors, such as smartwatches~\cite{laput2019sensing,wang2024washhandsbetteraccurate}, smartphones~\cite{sun2024multimodal,devrio2023smartposer}, earbuds~\cite{prakash2020earsense,zhao2024ui}, smart glasses~\cite{lee2018interaction,xie2021acoustic}, and Wi-Fi signals~\cite{ma2018signfi,wang2016wifall,wang2017device}. With accurate HAR, systems can anticipate user needs, improve safety, and provide context-aware assistance, fostering a seamless interaction between individuals and their surroundings.

In this paper, we take a step forward to propose a new HIAU task: action summarization using Wi-Fi and IMU signals in smart-home environments. As shown in Fig.~\ref{fig:fig1}, users perform continuous actions indoors, which are identified by recognizing each action’s category, start time, and end time to form a sequence of tuples. These tuples can be processed using large language models~(LLMs) to generate action summaries, such as insights into physical activity levels and behavioral patterns. With targeted prompts, the system can also detect anomalies like falls or missed medication. We envision several applications for action summarization: (1) Health Monitoring: providing detailed activity logs to support physical rehabilitation, track fitness goals, or detect health issues such as sedentary behavior or irregular movement patterns. (2) Elderly Care: detecting critical events like falls, identifying missed medication times, and offering behavior insights to caregivers for better elderly support. (3) Smart Home Automation: triggering adaptive responses, such as adjusting room temperature based on activity levels or controlling lighting when specific actions are performed. (4) Personal Assistants: Summarizing daily routines to help users with task planning, reminders, and prioritizing activities, thereby enhancing productivity.

The key technology enabling action summarization is temporal action localization (TAL), which focuses on accurately identifying and localizing the start and end times of actions within untrimmed sensory recordings. This step forms the foundation for generating tuples required for further summarization tasks. Over the years, numerous datasets have been developed to support TAL, primarily focusing on video-based tasks. However, these datasets are not well-suited for summarizing indoor daily activities due to several limitations: (1) Task-Specific Datasets: many datasets, such as YouCook2~\cite{zhou2018towards-youcook2}, Breakfast Dataset~\cite{kuehne2014language-breakfast}, and 50Salads~\cite{stein2013combining-50salads}, focus on specific tasks like cooking or food preparation. While these datasets are valuable within their domains, they lack diversity and fail to generalize to other types of everyday indoor activities. (2) Activity Types and Context: datasets like FineAction~\cite{liu2022fineaction}, THUMOS14~\cite{idrees2017thumos}, and ActivityNet~\cite{caba2015activitynet} predominantly feature sports or outdoor activities, which do not represent typical indoor scenarios encountered in daily life. (3) Privacy and Applicability: large-scale datasets, such as HACS Segment~\cite{zhao2019hacs} and ActivityNet~\cite{caba2015activitynet}, provide diverse activity coverage but are less applicable to indoor settings due to privacy concerns raised by cameras, which limit their use in monitoring indoor activities. In recent years, indoor-focused datasets have started to emerge. For instance: Ego-ADL~\cite{sun2024multimodal}: this dataset uses smartphone IMUs, Wi-Fi and audio for TAL but lacks key motion-sensing information from important body positions like the hands and head. This omission makes it challenging to differentiate fine-grained actions, and its reliance on audio raises significant privacy concerns for indoor monitoring applications. WiFiTAD~\cite{liu2024wificsibasedtemporal}: while it provides indoor TAL capabilities using Wi-Fi signals, its dataset includes only three participants in a single scene and seven actions, which limits its generalizability.

To address these limitations, we present XRF V2, a novel dataset designed specifically for summarizing indoor daily activities. XRF V2 integrates data from multiple sensing modalities, including Wi-Fi signals, IMU sensors (smartphones, smartwatches, headphones, and smart glasses), and synchronized video recordings, as illustrated in Fig.~\ref{fig:fig1}. XRF V2 combines Wi-Fi and IMU sensors to capture human motions, providing a more comprehensive understanding of actions. In real-world scenarios, sensor devices may experience malfunctions. For example, if a smartphone's IMU malfunctions or the device is not worn as intended (e.g., left on a table), the IMU data may no longer accurately represent the user's actions. Similarly, a malfunctioning Wi-Fi router can disrupt Wi-Fi signals. XRF V2 is designed to handle such situations by adapting to different combinations of sensor inputs for Temporal Action Localization (TAL) and action summarization. This flexibility ensures the system remains functional even when some sensors are not operating optimally. Moreover, Wi-Fi and IMU sensors are commonly found in commercial products, making it easy to integrate TAL and action summarization systems into existing home and personal devices without the need for costly specialized equipment.  This dataset was collected with the participation of 16 volunteers, who performed diverse and continuous action sequences across three distinct indoor environments: the dining room, study room, and bedroom, aiming to support a wide range of applications in action summarization using Wi-Fi and IMU signals in smart-home environments.  In total, XRF V2 consists of approximately 16 hours and 16 minutes of multimodal data recordings, comprising 853 annotated action sequences that encompass diverse daily routines, including sleeping, cutting fruit, and taking medicine. Compared to WiFiTAD~\cite{liu2024wificsibasedtemporal}, XRF V2 provides a more diverse, realistic, and comprehensive benchmark for the action summarization task.

We further propose the XRFMamba neural network to evaluate the XRF V2 dataset and conduct accurate Temporal Action Localization (TAL) for action summarization. XRFMamba is based on the latest advanced achievements in Mamba, which excels at capturing long-term dependencies in temporal sequences. We choose XRFMamba because, unlike action recognition, TAL involves longer input sequences that contain multiple actions, and Mamba has already demonstrated strong capabilities in handling long-term dependencies in fields like NLP~\cite{gu2023mamba} and CV~\cite{zhu2024vision}. For example, ActionMamba~\cite{chen2024video-mamba-suite} has been successfully applied for video-based TAL tasks. In this paper, we are the first to apply the Mamba network to handle Wi-Fi and IMU data for TAL tasks. Compared with existing methods such as ActionFormer~\cite{zhang2022actionformer}, TriDet~\cite{shi2023tridet}, ActionMamba~\cite{chen2024video-mamba-suite}, TemporalMaxer~\cite{tang2023temporalmaxer}, UWiFiAction~\cite{wang2023u}, and WiFiTAD~\cite{liu2024wificsibasedtemporal},  the proposed XRFMamba excels at capturing long-term dependencies in untrimmed sensory sequences and achieves the best performance with an average mAP of 78.74, outperforming the recent WiFiTAD by 5.49 points in mAP@avg while using 35\% fewer parameters. In action summarization, we introduce a new metric, Response Meaning Consistency (RMC), to evaluate action summarization. And it achieves an average Response Meaning Consistency(mRMC) of 0.802.

 Moreover, we conducted a leave-one-person-out evaluation, where XRFMamba achieved a mAP of 72.84 for new users, demonstrating strong generalization capabilities. We also performed device combination experiments to assess the model’s flexibility under different device configurations. Notably, even when using only a single right-hand-worn smartwatch, XRFMamba still achieved a mAP of 65.81, indicating that users can flexibly balance performance and device availability. Additionally, we tested XRFMamba's real-world deployment potential in a real apartment, where the system achieved an mAP of around 70, further validating its practical applicability.

Our contributions are threefold:

(1) XRF V2 Dataset: We have collected the XRF V2 dataset, which consists of 16 participants performing continuous indoor actions across three different environments, totaling 30 unique actions. This dataset provides a rich resource for indoor continuous action understanding tasks and systems, offering valuable data for advancing research in this area.

(2) New Task, Method, and Evaluation Metric: We introduce the novel task of action summarization using Wi-Fi and IMU signals in smart-home environments, a new human indoor action understanding task. We also propose a new evaluation metric, Response Meaning Consistency (RMC), to assess the performance of action summarization. We further design a novel method, XRFMamba, to effectively tackle this task. Compared to existing state-of-the-art methods, XRFMamba demonstrates superior performance in action summarization.

(3) Smart Home Intelligent Agents: We demonstrate that continuous sensory data from wearable and Wi-Fi devices, when combined with large language models (LLMs), can enable the development of intelligent agents within smart homes. These agents, such as personal assistants, health assistants, and home assistants, can offer various services, including task reminders, personalized health monitoring, climate control, and more, contributing to the intelligence of smart home systems.

\section{Related Work}\label{sec:related_work}

\subsection{Action Understanding with Wi-Fi Signals}\label{sec:wifi-action-understanding}

Action understanding with Wi-Fi signals has become a key technology in  human sensing. The RSSI (Received Signal Strength Indicator) in Wi-Fi signals is commonly used for indoor localization~\cite{bahl2000radar,wu2012will}. However, its accuracy is relatively low and is susceptible to multipath effects and temporal dynamics~\cite{yang2013rssi,wu2011chip}. As a result, later works have adopted more information-rich Channel State Information (CSI) to achieve more precise localization~\cite{yang2013rssi}, such as FILA~\cite{wu2012fila}, SpotFi~\cite{kotaru2015spotfi}, Splicer~\cite{xie2015precise}, and Dynamic-MUSIC~\cite{li2016dynamic}. In addition to high-precision indoor localization, Wi-Fi CSI signals have been widely applied in various tasks in fields such as smart homes, health monitoring, and human-computer interaction. These include hand gesture recognition~\cite{pu2013whole,ma2018signfi}, finger gesture recognition~\cite{li2016wifinger,tan2016wifinger,ali2015keystroke}, activity recognition~\cite{wang2015understanding,wang2016wifall,wang2014eyes}, vital sign monitoring~\cite{wang2016human,wang2017tensorbeat,yu2021wifi}, pose estimation~\cite{jiang2020towards,wang2019person,yan2024person,wang2019can}, mesh reconstruction~\cite{wang2022wi,wang2022wifi}, crowd counting~\cite{xi2014electronic}, and human imaging~\cite{xu2024wicamera,wang2019person}. Since Wi-Fi-based human sensing is highly sensitive to environmental changes, several studies have concentrated on enhancing its cross-environment capabilities. For example, EI~\cite{jiang2018towards} and Person-in-WiFi~\cite{wang2019person} use domain adaptation techniques to align the features of the source and target environments for activity recognition and pose estimation, respectively. CrossSense~\cite{zhang2018crosssense} proposes using a mixture of experts approach to train multiple tasks and select the optimal expert for the target environment, maintaining cross-environment adaptability. Widar 3.0~\cite{zheng2019zero}, on the other hand, proposes environment-invariant representation, i.e., body velocity profiles, to achieve cross-environment, cross-person, and cross-method hand gesture recognition.

Apart from the tasks mentioned above, some researchers have also focused on understanding continuous action using Wi-Fi. Smokey~\cite{zheng2016smokey} recognizes smoking actions by detecting periodic patterns, such as holding the cigarette, bringing it to the mouth, and inhaling the smoke. WiLife~\cite{li2024wilife} uses Wi-Fi for continuous in-home monitoring of elderly individuals living alone, ensuring their safety and well-being. 
Wang et al.~\cite{wang2023u} employ U-shaped deep neural networks for the temporal localization and segmentation of daily activities and human-human interactions. WiFiTAD~\cite{liu2024wificsibasedtemporal} introduces a frequency-aware learning approach combined with a dual-pyramid network to promote the accuracy of temporal action localization. IMar~\cite{he2022imar} employs tensor decomposition to extract individual Wi-Fi streams for each person, enabling separate recognition of continuous actions. In comparison to these works, we push the boundaries further by introducing a new task: action summarization using Wi-Fi and IMU signals in smart-home environments. We envision this task supporting applications in remote healthcare and monitoring, personal assistants, smart home automation, and more.

\subsection{Action Understanding with IMUs}\label{sec:imu-action-understanding}

Inertial Measurement Units~(IMUs), such as accelerometers and gyroscopes, can measure motion acceleration and angular velocity. They are widely embedded in personal consumer electronics, including smartphones, smartwatches, smart glasses, and earbuds. Researchers have leveraged IMUs in these devices to develop numerous applications in areas such as healthcare ~\cite{alam2020ai,jung2022imu,devrio2023smartposer,shang2024otago,wang2024washhandsbetteraccurate,de2020machine,wang2022social,meng2022mask}, sports~\cite{tian2021wearable,wolff2018activity}, human-computer interaction~\cite{laput2019sensing,devrio2023smartposer,sun2024multimodal,devrio2023smartposer,wahl2015using,scholl2015wearables-Wetlab,verma2021expressear,zhao2024ui}, and industrial applications ~\cite{azadi2019feasibility,sopidis2022micro}. Jung et al.~\cite{jung2022imu} utilize smartwatches and continuous wavelet transform (CWT) technology to identify tremor episodes and amplitude of Parkinson's disease patients during drawing tasks. 
Azadi et al.~\cite{azadi2019feasibility} investigate the feasibility of recognizing screwing actions in manufacturing using smartwatches and unsupervised learning techniques.
De et al.~\cite{de2020machine} place a mobile phone on a patient's left hip to estimate hip and knee joint loading, providing valuable insights for doctors in diagnosing the patient's condition.
Anguita et al.~\cite{anguita2013public-sbhar} leverage a smartphone worn at the waist to identify six daily activities: standing, sitting, lying down, walking, walking downstairs, and walking upstairs.
WISEGlass~\cite{wahl2015using} captures the wearer's head movements—up, right, left, and down—using IMUs embedded in the legs of smart glasses, enabling the control of the Pac-Man game accordingly.
Wetlab~\cite{scholl2015wearables-Wetlab} integrates Google Glass with a wrist-worn IMU sensor, allowing for the capture and review of DNA extraction experiments conducted in wet laboratory environments.
ExpressEar~\cite{verma2021expressear} utilizes a Nokia eSense earbuds to sense fine facial muscle movements, and mine the correlation between the movements and facial expressions for hands-free control using facial expressions or to monitor involuntary facial expressions to infer the user's emotional state.
Ui-Ear~\cite{zhao2024ui} utilizes earbuds to detect on-face gestures, including tapping, slide up, slide down, slide left, slide right, clockwise circle, and counterclockwise circle, enhancing the user-earbuds interaction experience.

Current advancements primarily focus on Human Activity Recognition (HAR), but a recent work~\cite{bock2024temporal} shifts its focus to Temporal Action Localization (TAL) using IMU data. It adapts video-based TAL methods, e.g., ActionFormer~\cite{zhang2022actionformer}, TemporalMaxer~\cite{tang2023temporalmaxer}, and TriDet~\cite{shi2023tridet}  on multiple public datasets, including WetLab for DNA
extraction experiment~\cite{scholl2015wearables-Wetlab} and Hang-Time HAR for basketball-playing actions~\cite{hoelzemann2023hang}. We take a step further by collecting a continuous dataset of indoor daily activities, specifically designed to evaluate both Temporal Action Localization (TAL) and action summarization tasks.

\subsection{Continuous Action Dataset and Benchmark}\label{sec:continuous-action-dataset}

\begin{table}[ht]
\small
\centering
\caption{XRF V2 shows great advantages in the number of scenes, actions, subjects, sequences, and modalities compared to the very latest WiFiTAD~\cite{liu2024wificsibasedtemporal}.}
\begin{tabular}{lllllll}
\toprule
Dataset      & \#Scene  & \#Actions  & \#Subjects & \#Sequences    & Modality  & Venue     \\
\midrule
WiFiTAD~\cite{liu2024wificsibasedtemporal}           & 1    & 7        & 3     & 553  & Wi-Fi & AAAI 2025  \\
XRF V2      & 3 & 30 & 15 & 853 (16h 16m 8s)                                      & Wi-Fi, 5 IMUs, RGB+D+IR & IMWUT 2025\\ 
\bottomrule 
\end{tabular}
\label{tab:dataset}
\end{table}

Temporal action localization~(TAL) and temporal action detection~(TAD) are fundamental tasks in the computer vision community and have been extensively studied~\cite{shou2016temporal}. Searchers have introduced numerous publicly accessible datasets to evaluate  TAL and TAD methods. Breakfast~\cite{kuehne2014language-breakfast} and YouCook2~\cite{zhou2018towards-youcook2} are two representative large-scale daily cooking datasets captured from third-person viewpoint cameras. Beyond these cooking datasets, some other datasets such as ActivityNet~\cite{caba2015activitynet}, THUMOS14~\cite{idrees2017thumos}, HACS Segments~\cite{zhao2019hacs},  FineAction~\cite{liu2022fineaction}, are released for TAL involving daily activities, sports, and household tasks. UWash~\cite{wang2024washhandsbetteraccurate} is a dedicated system for segmenting handwashing gestures using smartwatch IMU data. Ego-ADL~\cite{sun2024multimodal} leverages a smartphone to continuously capture users' audio, Wi-Fi, and motion sensor data in daily life. WiFiTAD~\cite{liu2024wificsibasedtemporal} uses a setup with one transmitter and one receiver Wi-Fi device to collect data in a single indoor environment. 

Compared to UWash, Ego-ADL, and WiFiTAD, XRF V2 offers richer sensor data, integrating two smartphones, two smartwatches, one pair of earbuds, and one smart glasses. This comprehensive setup captures diverse motion information, including body, hand, and head movements, enabling more effective Temporal Action Localization (TAL) in home environments. Since WiFiTAD is the most relevant and latest work to XRF V2, we compare XRF V2 with it in Table~\ref{tab:dataset}, which demonstrates significant advantages over WiFiTAD in terms of the number of scenes, actions, subjects, sequences, and modalities. Furthermore, XRF V2 supports the exploration of optimal sensor combinations and multimodal data fusion, providing a comprehensive dataset for advancing IMU and Wi-Fi based temporal action localization and action summarization research.

\section{XRF V2 Acquisition and Description}\label{sec:XRF V2}
\subsection{Hardware Setups}\label{sec:wifi-imu-devices}

\textbf{Wi-Fi transceivers.} We used ThinkPad X201 laptops equipped with Intel 5300 wireless network cards as the Wi-Fi transceivers. The Wi-Fi system consists of one transmitter and three receivers, positioned at the four corners of a 3.1 m$\times$2.4 m rectangular area. This setup is inspired by deployment strategies used in several studies, such as Widar3.0~\cite{zheng2019zero}, OneFi~\cite{10.1145/3485730.3485936}, and XRF55~\cite{wang2024xrf55}, which have shown that a multi-device configuration like this can enhance sensing performance. The devices are placed at a height of 120 cm to ensure effective coverage for full-body motion capture~\cite{8740950}. The transmitter operates with a single antenna, broadcasting packets at 200  per second on channel 128 (5.64 GHz). Each receiver is equipped with three antennas. Using the Wi-Fi CSI Tool~\cite{halperin2011tool}, we extract the channel state information (CSI) from 30 orthogonal frequency-division multiplexing (OFDM) subcarriers. The resulting CSI recordings are represented as a tensor of size $(200t) \times 1 \times 3 \times 3 \times 30$, where $t$ is the recording duration in seconds. For example, for an 80-second continuous action, the CSI recordings yield a tensor of size $16000 \times 1 \times 3 \times 3 \times 30$, capturing detailed variations in the wireless channel over time for subsequent analysis.

\textbf{IMU devices.} The IMU device choice follows IMUPoser~\cite{mollyn2023imuposer} and includes three commonly used devices in daily life: smartwatches, smartphones, and earbuds. Additionally, we introduce smart glasses, considering their promising potential in AR and the metaverse~\cite{zhao2023heads}. In our experiments, the brand of earbuds is AirPods Pro, with IMU data recorded at a sampling rate of 25 Hz. For smartwatches, smartphones, and smart glasses, we use commercially available IMU modules to simulate the IMUs of these devices. These modules allow centralized control through a PC-based interface, facilitating better synchronization across all IMU devices. The IMU modules record data at a sampling rate of 50 Hz. The placement of IMU devices is shown in Fig.~\ref{fig:fig1}.

\textbf{Azure Kinect:} The Azure Kinect captures RGB, depth, and infrared images at 15 frames per second in 720P resolution using its toolkit. The Kinect is placed between the transmitter and the third receiver at a height of 115 cm to capture video footage of the movement of individuals in the scene. We can use open-source computer vision models to extract pose and mesh information from the videos, which will be useful for future tasks such as IMU- and Wi-Fi-based pose estimation and mesh reconstruction.

\begin{figure}[t]
    \centering
    \includegraphics[width=1\linewidth]{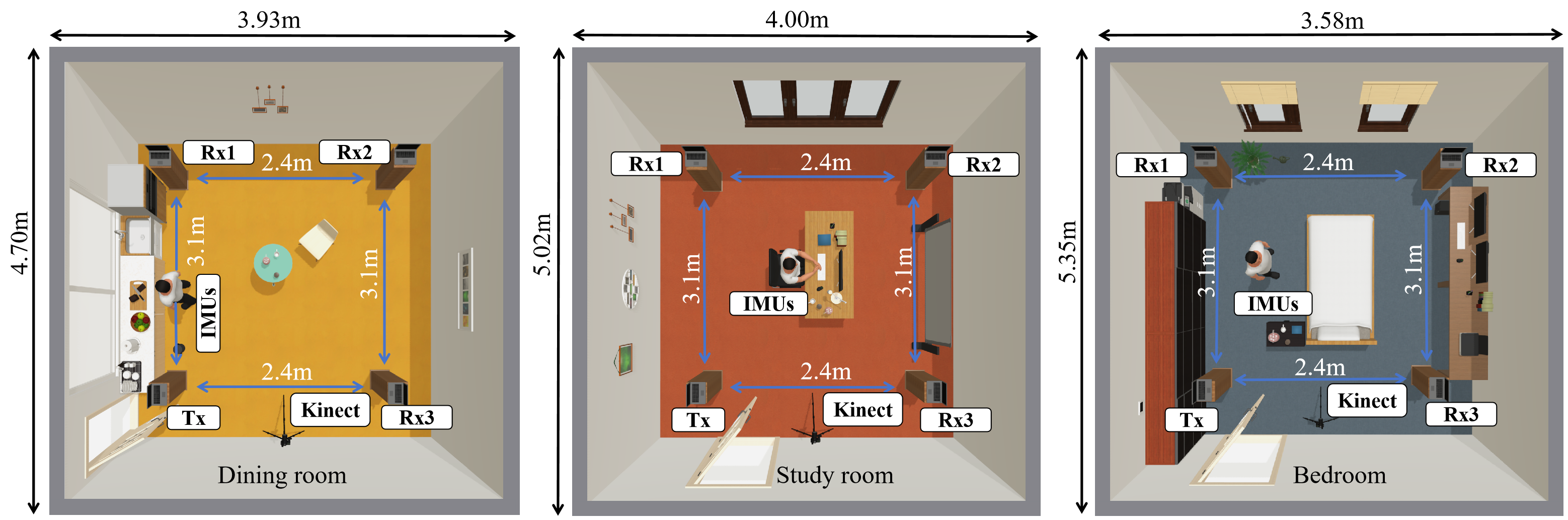}
    \caption{Illustration of the experimental scenarios, where volunteers perform continuous action sequences in the dining room, study room, and bedroom. IMU devices from two smartwatches, two smartphones, a pair of earbuds, a pair of glasses, and four Wi-Fi transceivers, along with an Azure Kinect, are synchronized to record action sequences.}
    \label{fig:scene}
\end{figure}

The experimental scenarios are shown in Fig.~\ref{fig:scene}. Before data recording, all the aforementioned hardware devices undergo careful synchronization. Notably, XRF V2 selects commonly available consumer-grade products used in everyday household and personal settings, unlike XRF55~\cite{wang2024xrf55}, which utilizes millimeter-wave radar and an RFID reader with a tag array.

\begin{table}[t]
\centering
\caption{Actions proposed and executed by volunteers in the Bedroom, Study room, and Dining room.}
\begin{tabular}{rp{13cm}}
\toprule 
\textbf{Scene} & \textbf{Proposed actions}  \\
\midrule  
Dining room        & 1. Walk (towards the cabinet, towards the chair), 2. Sit down, 3. Stand up, 4. Pour water into the cup, 5. Drink water, 6. Take medicine, 7. Pick up things, 8. Take the fruits from the cabinet, 9. Cut fruits, 10. Eat fruits, 11. Wash hands, 12. Throw waste, 13. Wipe the table, 14. Stretch when sitting. \\
\midrule 
Study room         & 1. Walk (towards the door, towards the chair, towards the blackboard), 2. Sit down, 3. Stand up, 4. Pour water into the cup, 5. Drink water, 6. Take medicine, 7. Turn on and off the desk lamp, 8. Operate the mouse, 9. Write, 10. Operate the keyboard, 11. Read a book, 12. Open an envelope, 13. Answer the phone, 14. Stretch when standing, 15. Write on the blackboard. \\
\midrule  
Bedroom        &  1. Walk (towards the bed, towards the window), 2. Get up, 3. Sit down, 4. Lie down, 5. Stand up, 6. Read a book, 7. Pour water into the cup, 8. Drink water, 9. Take medicine, 10. Stretch when standing, 11. Use phone, 12. Open and close windows, 13. Open and close curtains, 14. Water plants, 15. Stand still, 16. Lying still. \\
\bottomrule  
\end{tabular} 
\label{tab:proposed-action}
\end{table}

\subsection{Action Sequence Proposal}\label{sec:action-sequence-proposal}

We select three typical household scenarios for the experiment: the dining room, study room, and bedroom. These settings allow us to summarize users' activities related to eating, studying, and sleeping, along with an associated series of actions. For example, activities such as taking medicine, drinking water, washing hands, and stretching serve as health-related indicators; and actions like opening curtains or turning on lights provide insights into interactions with household devices.

We communicate the above objective to the volunteers. Instead of assigning each person a fixed action sequence, we aim to increase the diversity of action sequences and incorporate personal habits relevant to the volunteers. Therefore, we ask each volunteer to recall their own daily action sequences in the dining room, study room, and bedroom, and estimate the execution time of each action in the sequence. Each volunteer recalls five action sequences based on the starting positions of the actions, with each sequence containing 7-10 actions. Fig.~\ref{fig:action-sequences} shows examples of some action sequences, while Table~\ref{tab:proposed-action} presents the proposed actions in each scene. In Appendix~\ref{sec:action-descript-and-value}, we demonstrate that these actions can be integrated with a large language model (LLM) assistant to enhance smart home intelligence.

\subsection{Volunteer Coordination and Action Execution}\label{sec:volunteer-coordination}

We recruit 16 volunteers, with an age range of 22–34 years, a height range of 1.57–1.82 meters, and a weight range of 43–90 kgs. Volunteers can access the informed consent form online, which includes details about the study's purpose, methods, participation duration, potential risks, and compensation.

Before performing the action sequences, we proportionally shorten the durations estimated by the volunteers to fall within 5–20 seconds. This adjustment is necessary because the originally estimated durations are often much longer; for instance, reading a book might be estimated as one hour. Performing such lengthy actions would significantly increase data collection time and complicate annotation. We also make reasonable adjustments to the execution durations several times, generating more action sequences. As a result, the action sequences in each scene become more diverse. 

After that, we can adopt an automated annotation method to streamline the process. The action sequences proposed by the volunteers are converted into audio prompts using text-to-audio technology. These audio prompts are played in real time to guide the volunteers. Upon hearing an action instruction, the volunteer performs the corresponding action and transitions to the next action upon hearing the subsequent prompt. This process continues until the entire sequence is completed.

\begin{figure}[ht]
    \centering
    \includegraphics[width=\linewidth]{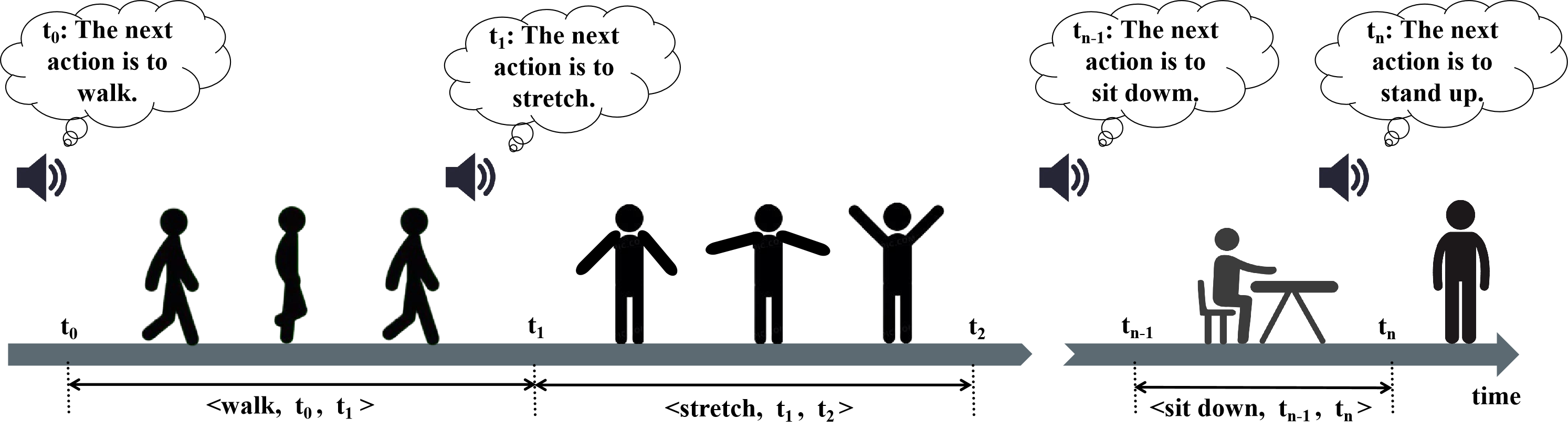}
    \caption{Automatic annotation process. The audio prompts are played in real-time to guide the volunteers. Upon hearing an action instruction, the volunteer performs the corresponding action and transitions to the next action upon hearing the subsequent prompt. The start time and end time of an action are annotated with the prompts played time.}
    \label{fig:automatic-annotating}
\end{figure}

\begin{figure}[t]
    \centering
    \includegraphics[width=1\linewidth]{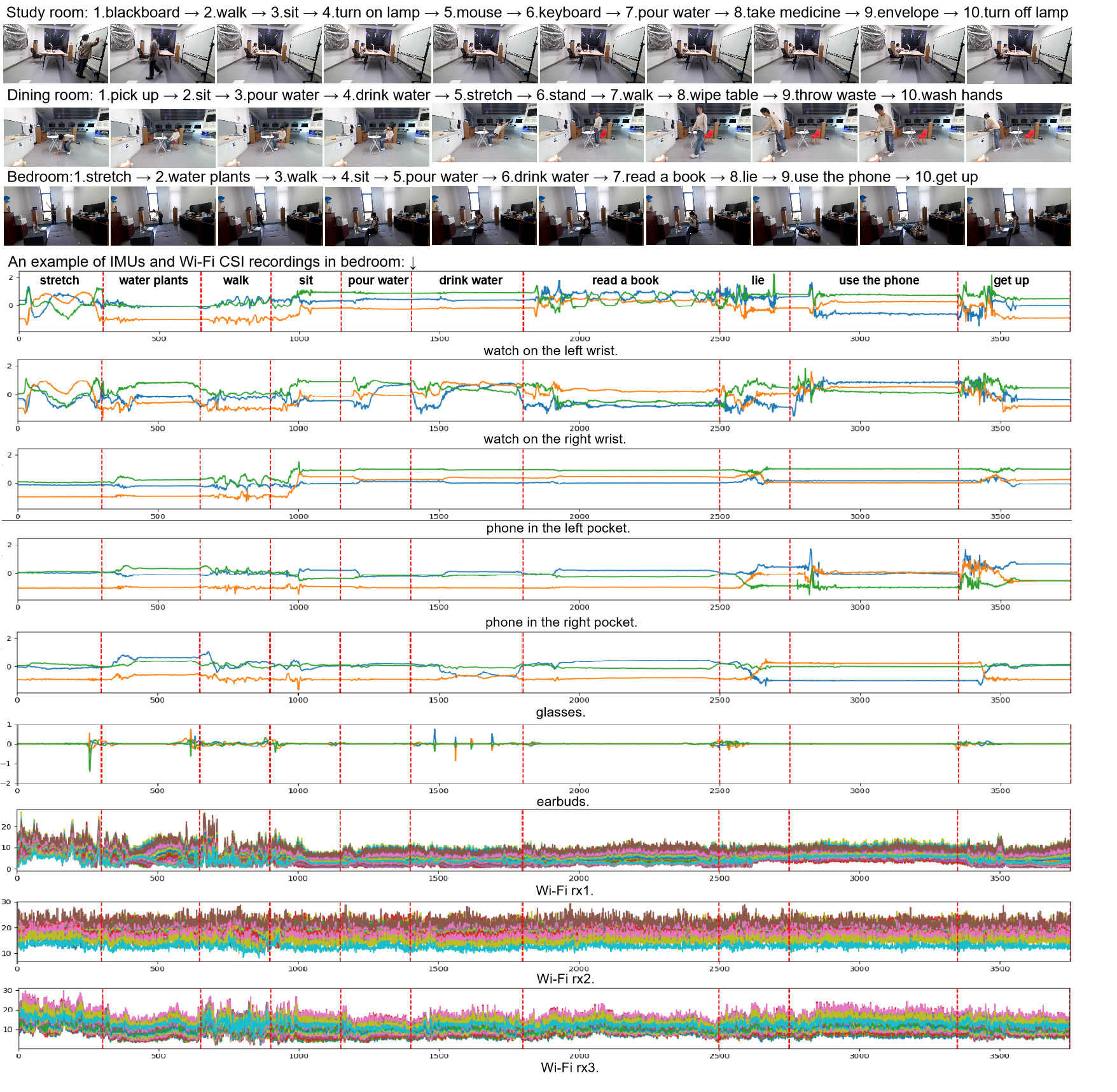}
    \caption{Examples of action sequence from the study room, dining room, and bedroom, with a detailed visualization of the sensory sequence from the bedroom.}
    \label{fig:action-sequences}
\end{figure}

The start and end times of each audio prompt are automatically recorded as the start and end times of the corresponding action. Fig.~\ref{fig:automatic-annotating} illustrates this automatic annotation process. As a result, after executing an action sequence containing $n$ actions, we automatically obtain the annotations for the entire sequence: 
$\{(a_i, t_i^s, t_i^e), i = 1, 2, \dots, n\}$,
where $a_i$, $t_i^s$, and $t_i^e$ represent the action category, start time, and end time of the $i$-th action, respectively. 

In general, the duration of a single action sequence is approximately 50--80 seconds. After completing one action sequence, volunteers rest for 10 seconds, and after executing five sequences, they take a 5-minute break. These brief breaks provide the volunteers with proper rest, reduce fatigue, and maintain the overall quality of the data collection process. It also allows us to check the functionality of various devices and minimizes the cost of re-collecting data.

In Fig.~\ref{fig:action-sequences}, we present action sequences from the study room, dining room, and bedroom, with a sensory sequences visualization from the bedroom. For clarity, we display only the three-axis accelerometer data from all IMU devices, and for the Wi-Fi data, we select the data from one antenna of each receiving device. Additionally, the sampling frequency of all data has been standardized, and the x-axis represents the timestamp. To highlight the different actions in the sequence, we use dashed lines in the figure to indicate the endpoints of each action.

\subsection{Dataset Statistics}\label{sec:dataset-statistics}

Each of the 16 volunteers proposed 5 action sequences for each scene. By shortening durations and adjusting the length of individual actions, we generated multiple variants of these sequences. Specifically, in the bedroom, each volunteer performed three different variations of each action sequence, while in the study room and dining room, four different variations were performed for each sequence. Thus, XRF V2 dataset contains a total of: $16 \times 5 \times 3 + 16 \times 5 \times 4 + 16 \times 5 \times 4 = 880$ action sequences.  After filtering out incomplete and erroneous sequences, we obtained 853 valid action sequences, with a total duration of 16 hours, 16 minutes and 8 seconds. These sequences were captured using synchronized Channel State Information (CSI) from four Wi-Fi transceivers, IMU data from two smartwatches, two smartphones, a pair of earbuds, and a pair of smart glasses, along with RGB+D+IR video streams from a Kinect camera.
Furthermore, we implemented a stratified partitioning strategy maintaining an 80:20 ratio between training and testing sets across all environmental contexts and participants, resulting in 682 sequences (80\%) allocated for model training and 171 sequences (20\%) reserved for evaluation purposes. Table~\ref{tab:statistics} shows detailed statistics of XRF V2 dataset.

\begin{table}[ht]
\caption{XRF V2 contains 853 multimodal sensory sequences for action summarization and temporal action localization.}
\begin{tabular}{lllll}
\hline
            & \multicolumn{2}{c}{train} & \multicolumn{2}{c}{test} \\ \hline
Dining room & 244 sequences & 4h39min41s & 61 sequences & 1h8min40s \\
Study room  & 256 sequences & 5h0min31s  & 64 sequences & 1h16min15s \\
Bedroom     & 182 sequences & 3h19min8s  & 46 sequences & 51min53s  \\ \hline
Total       & 682 sequences & 12h59min20s & 171 sequences & 3h16min48s \\ \hline
\end{tabular}
\label{tab:statistics}
\end{table}

\section{Background}\label{sec:background}

State Space Models (SSMs), such as S4~\cite{gu2021efficiently}, S5~\cite{smith2022simplified}, and Mamba~\cite{gu2023mamba}, are a recent class of deep learning methods that have demonstrated advantages in both natural language processing~\cite{gu2023mamba} and computer vision~\cite{zhu2024vision}. Unlike Transformer-based models, which suffer from quadratic time complexity due to self-attention, SSMs operate with linear-time complexity, making them significantly more efficient for long sequence modeling. In action summarization, the input sensor sequences are much longer than those used in action recognition tasks, necessitating the ability to effectively model long-range dependencies. To this end, we base our approach on an advanced SSM model and propose the XRFMamba network.

\begin{figure}[ht]
    \centering
    \includegraphics[width=1\linewidth]{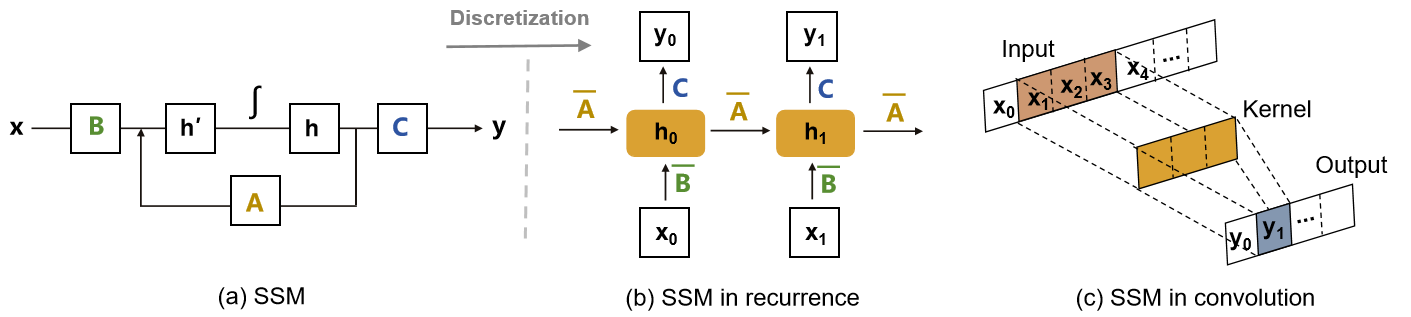}
    \caption{State space model (a). The model can be computed in linear recurrence (b) for inference, or global convolution (c) for training. }
    \label{fig:ssm}
\end{figure}

As introduced in Mamba~\cite{gu2023mamba}, state space models are inspired by a continuous system, that maps 1-dimensional function or sequence  $x(t) \in \mathbb{R} \mapsto y(t) \in \mathbb{R}$ through a hidden state $h(t) \in \mathbb{R}^\mathtt{N}$, illustrated in Fig.~\ref{fig:ssm} (a). The system works as follows:
\begin{equation}
\label{eq:ssm-continuous}
h'(t)  = \mathbf{A}h(t) + \mathbf{B}x(t), \hspace{10pt} 
y(t)  = \mathbf{C}h(t).
\end{equation}
where $\mathbf{A} \in \mathbb{R}^{\mathtt{N} \times \mathtt{N}}$ is the evolution parameter, and $\mathbf{B} \in \mathbb{R}^{\mathtt{N} \times 1}$ and $\mathbf{C} \in \mathbb{R}^{1 \times \mathtt{N}}$ are the projection parameters.

For practical implementation, the continuous system needs to be discretized. When discretizing, the continuous state transition matrices $(\mathbf{A}, \mathbf{B})$ are approximated by their discrete counterparts $(\mathbf{\overline{A}}$, $\mathbf{\overline{B}})$. The Zero-Order Hold~(ZOH) is commonly used in the  discretization process, defined as follows:
\begin{equation}
\label{eq:zoh}
\mathbf{\overline{A}} = \exp{(\Delta\mathbf{A})}, \hspace{10pt} 
\mathbf{\overline{B}} = (\Delta \mathbf{A})^{-1}(\exp{(\Delta \mathbf{A})} - \mathbf{I}) \cdot \Delta \mathbf{B}.
\end{equation}

After the discretization, Equation~(\ref{eq:ssm-continuous}) can be rewritten as:
\begin{equation}
\label{eq:discrete_lti}
h_t = \mathbf{\overline{A}}h_{t-1} + \mathbf{\overline{B}}x_{t}, \hspace{10pt}
y_t = \mathbf{C}h_t.
\end{equation}
Since this is in the linear recurrence expression, as illustrated in Fig.~\ref{fig:ssm} (b),  it is efficient for autoregressive inference.

In addition, the discretized SSM can also be expressed in the global convolution expression, as follows: 
\begin{equation}
\label{eq:conv}
\mathbf{\overline{K}} = (\mathbf{C}\mathbf{\overline{B}},  \mathbf{C}\mathbf{\overline{A}}\mathbf{\overline{B}}, \dots, \mathbf{C}\mathbf{\overline{A}}^{\mathtt{L}-1}\mathbf{\overline{B}}), \hspace{10pt}
\mathbf{y} = \mathbf{x} * \mathbf{\overline{K}},
\end{equation}
where $\mathtt{L}$ is input sequence length, and $\overline{\mathbf{K}} \in \mathbb{R}^{\mathtt{L}}$ is a structured convolutional kernel, illustrated in Fig.~\ref{fig:ssm} (c). The convolution expression is efficient for parallel training.

\begin{figure}[ht]
    \centering
    \includegraphics[width=0.5\linewidth]{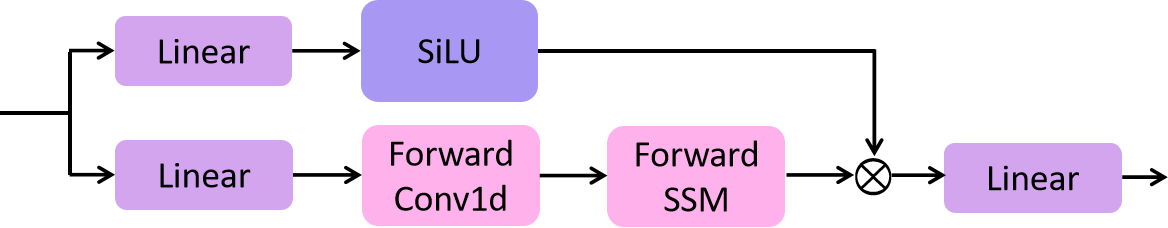}
    \caption{Basic Mamba Block~\cite{gu2023mamba}. }
    \label{fig:basic-mamba-block}
\end{figure}

Due to the flexibility of SSMs, they can switch to convolution during training and to recurrence during inference, providing fast performance in both stages. SSMs have a rich foundational knowledge, and introducing all aspects is beyond the scope of this paper. For more detailed information, please refer to Mamba~\cite{gu2023mamba} (structure shown in Fig.~\ref{fig:basic-mamba-block}) . In practice, SSMs are highly convenient. For example, in Vision Mamba~\cite{zhu2024vision}, the Mamba module replaces the Transformer module in the Vision Transformer (ViT)~\cite{dosovitskiy2021imageworth16x16words}, achieving better performance than ViT while also being more computationally efficient. Additionally, in the Video Mamba Suite~\cite{chen2024video-mamba-suite}, Mamba is evaluated as a plug-in or backbone, showing excellent results across 12 video understanding tasks, including action recognition and temporal action localization.

\section{Method}\label{sec:method}

\subsection{XRFMamba}\label{sec:xrfmamba}

\begin{figure}[ht]
    \centering
    \includegraphics[width=1\linewidth]{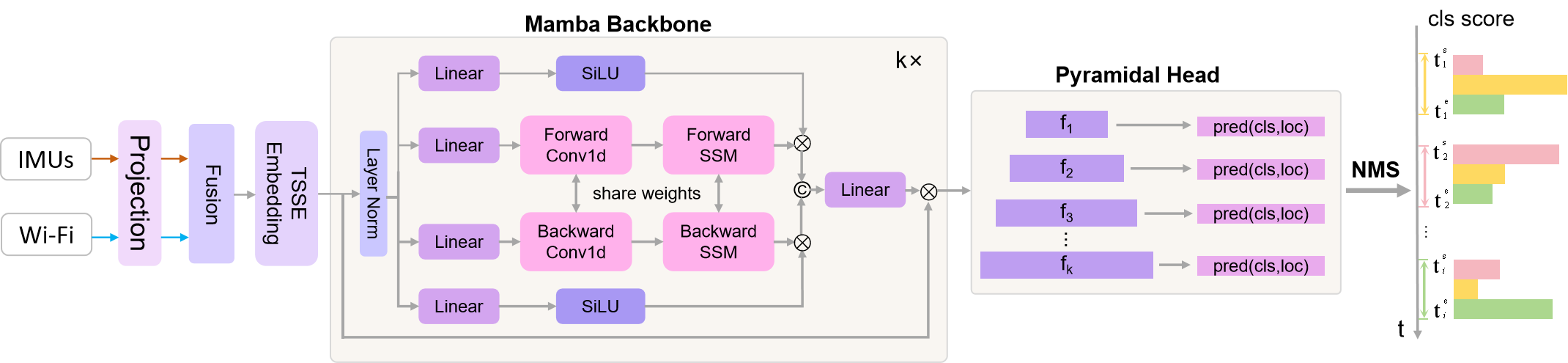}
    \caption{XRFMamba uses Mamba as the feature learning backbone. It takes an input sequence and predicts the action categories that appear in the sequence, along with the start and end times of each action. }
    \label{fig:xrfmamba}
\end{figure}

To achieve action summarization, it is essential to identify the actions performed by the user within a specific time span, along with their respective start and end times, which corresponds to the Temporal Action Localization (TAL) task. To address TAL, we propose XRFMamba, a novel neural network based on the State Space Model (SSM). As shown in Fig.~\ref{fig:xrfmamba}, XRFMamba employs Mamba blocks as the representation learning backbone, enabling the accurate localization of actions in continuous sensory sequences. 

\textbf{(1) Projection and Fusion}. The Projection module is adapted from WiFiTAD~\cite{liu2024wificsibasedtemporal} and comprises three sequential components: 1D Convolutional Operations, Group Normalization, and ReLU Activation. The Projection module standardizes the output channels of IMU and Wi-Fi data using convolutional operations, ensuring consistent channel dimensions for the subsequent fusion operation. The fusion of Wi-Fi and IMU data is performed through a weighted summation, as shown in the following equation:
\begin{equation}
    e = \lambda e_{wifi} + (1-\lambda)e_{imu}
\end{equation}
Here, $e_{wifi}$ and  $e_{imu}$ represent Wi-Fi and IMU data outputs after projection, respectively. In our experiments, we observed that the IMU data performed better than Wi-Fi data. To maintain the stability of the fusion performance, we assigned a higher weight to the IMU data. Specifically, we set $\lambda=0.2$ in our experiments. In this stage, the Projection module primarily ensures that the channel dimensions of the two data sources are consistent, preparing them for the subsequent fusion operation. The fusion module effectively combines the information from Wi-Fi and IMU, which is subsequently fed into further processing.

\textbf{(2) Embedding}. After fusion, the design of our embedding module takes inspiration from ActionMamba~\cite{chen2024video-mamba-suite}. In ActionMamba, the input to the Mamba backbone is not raw video but feature embeddings derived from the pre-trained model, InterVideo~\cite{Internvideo}. Following a similar approach, we introduce an embedding step prior to feeding the data into the Mamba backbone. However, due to the absence of widely accepted pre-trained models for IMU and Wi-Fi data in both academic and industrial contexts, a dedicated module is required to generate these embeddings. In this regard, we leverage the TSSE (Temporal Signal Semantic Embedding) module from WiFiTAD, which has proven effective in embedding Wi-Fi data for temporal action localization tasks. The TSSE embedding module is a dual-stream architecture, where one stream is based on the Transformer, which excels at learning global features, and the other is based on convolutional layers, which are effective at capturing local features. These two streams are fused at the output stage to create a comprehensive feature representation. The fused embeddings are then passed into the Mamba backbone for further processing. The TSSE module learns the basic representations for both IMU and Wi-Fi data, overcoming the challenge of preparing representations for the Mamba backbone in the absence of pre-trained models for these modalities.

\textbf{(3) Mamba Backbone}. The Mamba backbone in Fig.\ref{fig:xrfmamba} is the Decomposed Bidirectionally Mamba (DBM), introduced in ActionMamba~\cite{chen2024video-mamba-suite}. DBM extends the original Vision Mamba (ViM)~\cite{zhu2024vision} and vanilla Mamba~\cite{gu2023mamba}. We present DBM block details in Algorithm.~\ref{alg:dbm}~(Appendix~\ref{sec:dbm}), consistent with the ViM algorithm style. Initially, the input sensory sampling point $\mathbf{S}_{l-1}$ is normalized and processed through four linear layers to produce $\mathbf{z_1}$, $\mathbf{x_{forward}}$, $\mathbf{x_{backward}}$, and $\mathbf{z_2}$. $\mathbf{x_{forward}}$ and $\mathbf{x_{backward}}$ are then processed in their respective forward and backward directions. Specifically, $\mathbf{x_{forward}}$ undergoes forward processing to yield $\mathbf{x'}_{forward}$, $\mathbf{B}_{forward}$, and $\mathbf{C}_{forward}$, which are subsequently discretized into $\overline{\mathbf{A}}_{forward}$ and $\overline{\mathbf{B}}_{forward}$. Next, $\mathbf{y_{forward}}$ is computed through the forward SSM and fused with $\mathbf{z_1}$ to generate $\mathbf{y_{forward}^{'}}$. Similarly, $\mathbf{y_{backward}^{'}}$ is obtained in the backward direction. The forward and backward outputs, $\mathbf{y_{forward}^{'}}$ and $\mathbf{y_{backward}^{'}}$, are then fused to produce the final output $\mathbf{S}_{l}$. Notably, the convolution and SSM parameters are shared between both directions in DBM. The adoption of the Mamba backbone offers significant advantages for temporal action localization based on Wi-Fi and IMU data. Its capability to efficiently model long-range dependencies, combined with a bidirectional architecture, enables more precise capture of temporal dynamics and accurate identification of action boundaries. Moreover, the linear computational complexity and parameter sharing across forward and backward paths make Mamba particularly well-suited for processing fused multimodal sensor data in a computationally efficient and scalable manner.

\textbf{(4) Pyramidal Prediction Head}. We use Mamba to transform the input into features with the same number of channels but different lengths, resulting in a feature pyramid composed of $(f_1, f_2, \dots, f_K)$. Each feature dimension $f_i$ is processed through two separate convolutional paths to output action recognition results and action localization results, respectively. The total loss is the sum of the errors from these two prediction branches. The overall training loss of the network is given as follows:
\begin{equation}
\label{eq:loss}
\begin{aligned}
 L^i &= \alpha_1 L_{Cls}^{i} + \alpha_2 L_{Loc}^{i} \\
   L &= \sum_{i=1}^K L^i
\end{aligned}
\end{equation}
where we use Focal Loss~\cite{lin2017focal} for action classification and $L_1$ loss for action localization; 
$\alpha_1$ and $\alpha_2$ are used to balance the scale of the two losses, and we set $\alpha_1 = 1$ and $\alpha_2 = 1000$ based on empirical observations to ensure that both losses are on a similar magnitude. This pyramidal prediction strategy leverages the ability of different lengths to capture actions of varying durations, originally applied in object detection~\cite{lin2017feature}, where features of different sizes capture objects of different scales, and has been widely adopted in temporal action localization tasks, such as in~\cite{zhang2022actionformer,liu2024wificsibasedtemporal,chen2024video-mamba-suite}.

\begin{figure}[ht]
    \centering
    \includegraphics[width=0.9\linewidth]{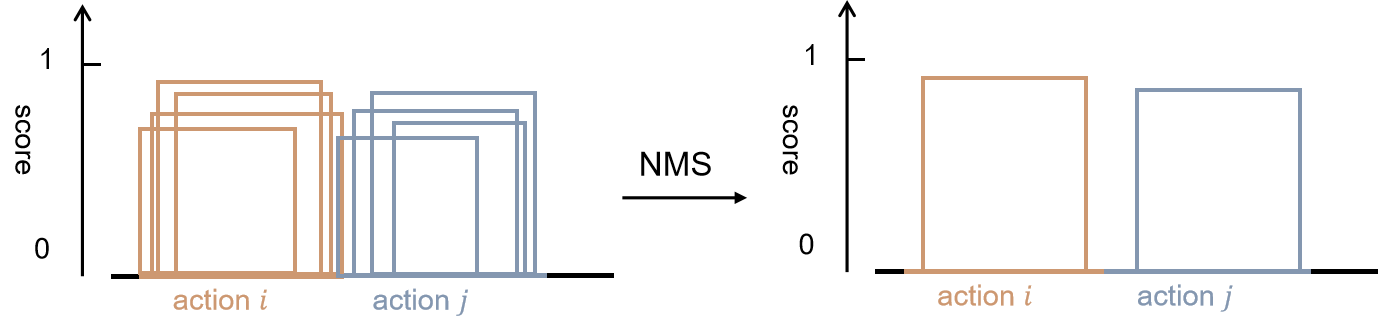}
    \caption{ Each action has multiple localization priors, leading to several candidate prediction results. NMS enables XRFMamba to keep only the best result for each action. (Illustration inspired by RF-fall~\cite{tian2018rf})}
    \label{fig:nms}
\end{figure}

(5) Inference. In XRFMamba, each action has multiple localization priors, leading to several candidate prediction results. To convert multiple localization prediction results into a single prediction, we use Non-Maximum Suppression (NMS). The steps of NMS are as follows: (1) Sorting: sort all candidate segments by their scores, processing the segment with the highest score first.
(2) Discard Overlapping Segments: after processing the current segment, compute its overlap with other segments. If the overlap exceeds a predefined threshold, that segment is discarded. (3) Repeat the Process: continue with the next highest-scoring segment and repeat step 3 until all segments have been processed.  Fig.~\ref{fig:nms} shows an example. The segment corresponding to the highest classification score for action $i$ is selected first, and the other segments that have a high overlap with it are discarded. As a result, action $i$ will only retain one localization prediction. Then, among the remaining segments, the segment with the highest score for action $j$ is selected, and all the other segments with a high overlap with it are discarded. Therefore, action $j$ will also have only one prediction result. NMS enables XRFMamba to keep only the best result for each action.

\subsection{Implementation Details}\label{sec:training-details}

Prior to training, we perform data augmentation on action sequence sensory data in the training set through sliding window segmentation. Each 30-second clip is generated with a 3-second sliding stride, expanding the original 682 training action sequences into 9660 30-second clips.  Subsequently, we unify multimodal sensor clips to 2048 temporal dimensions via signal interpolation: (1) Earbuds clips (25Hz$\times$30s = 750 samples) and other IMU clips (50Hz$\times$30s = 1500 samples)  are upsampled to 2048 using linear interpolation;  (2) Wi-Fi clips (200Hz$\times$30s = 6000 samples) are adaptively downsampled to 2048.
 
The sliding-window data augmentation introduces inherent challenges of partial action truncation, as 30-second clips contain multiple action instances. Temporal boundaries of clips frequently manifest partial action executions (head/tail truncation), where left boundaries preserve end time without start time, while right boundaries retain start time lacking end time. To address this truncation issue, we (1) incomplete segments covering $\ge80\%$ of original action duration are retained, with truncation points redefined as start/end time; (2) Segments containing $<80\% $ complete actions are excluded from loss computation via temporal masking during backpropagation.

We use AdamW optimizer to train XRFMamba for 80 epochs on one Nvidia 3090 GPU.  The initial learning rate is 4e-5, and is adjusted every 30 epochs by a factor of 0.5. The batch size is 8. 

During the inference phase, XRFMamba processes 30-second temporal clips with a 3-second sliding stride, generating preliminary action proposals for each clip. To consolidate overlapping predictions, we implement the temporal non-maximum suppression (NMS) algorithm described above. The final temporal action localization (TAL) outputs comprise action classifications with precise start/end time boundaries.

\section{Evaluation and Results}\label{sec:evaluation-results}

\subsection{Metrics} \label{sec:metrics}

\textbf{(1) AP and mAP for Temporal Action Localization.} We use the intersection over union (IoU) between the predicted action time and the ground truth time, referred to as tIoU~\cite{zhang2022actionformer,tang2023temporalmaxer,shi2023tridet}, to evaluate the performance of Temporal Action Localization (TAL). A higher tIoU indicates better performance. tIoU can be computed as follows:

\begin{equation}\label{eq:tiou}
tIoU = \frac{time~of~overlap}{time~of~union } = \frac{\max(0, \min(e_{\text{pred}}, e_{\text{gt}}) - \max(s_{\text{pred}}, s_{\text{gt}}))}{(e_{\text{pred}} - s_{\text{pred}}) + (e_{\text{gt}} - s_{\text{gt}}) - \max(0, \min(e_{\text{pred}}, e_{\text{gt}}) - \max(s_{\text{pred}}, s_{\text{gt}}))}
\end{equation}
where $s_{\text{pred}}$ and  $e_{\text{pred}}$ are the predicted start and end times of an action; $s_{\text{gt}}$ and  $e_{\text{gt}}$ are the ground truth start and end times. 

Furthermore, we use AP@t to denote the ratio of actions whose tIoU is greater than or equal to a threshold $t$, computed as follows:
\begin{equation}\label{eq:ap-tiou}
  AP@t = \frac{\sum_{i=1}^N \mathtt{I}( tIoU_i \ge t)}{N}
\end{equation}
where $N$ is the number of actions to compute tIoU. $\mathtt{I}$ outputs 1 if $IoU_i \ge t$ is true, else 0. The larger the AP@t, the higher the ratio of actions with a tIoU greater than or equal to the threshold $t$. On the other hand, as $t$ increases, the threshold for achieving a higher tIoU becomes more challenging, leading to a decrease in AP@t. The upper bound of AP@t is 1, while the lower bound is 0.

Besides, we use mAP (mean Average Precision) as the ultimate metric to evaluate the accuracy of Temporal Action Localization (TAL). It is computed by averaging the AP@t over 10 tIoU levels, specifically AP@[0.50:0.05:0.95], which covers tIoU thresholds from 0.5 to 0.95 with a step size of 0.05.

\begin{figure}[ht]
    \centering
    \includegraphics[width=1\linewidth]{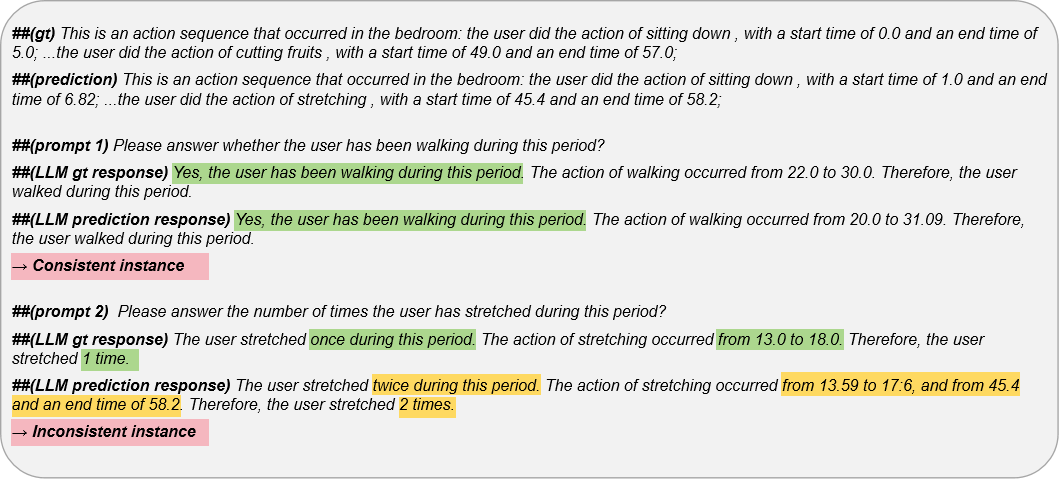}
    \caption{Example of Response Meaning Consistency (RMC) calculation. For each test action sequence, several actions with start and end times are provided for both ground truth and predicted values. Using a task-oriented prompt, the LLM generates corresponding action summarization results. Human auditors then assess whether the responses for the real and predicted values convey consistent meanings. A higher level of consistency results in a higher RMC score.}
    \label{fig:rmc}
\end{figure}

\textbf{(2) RMC and mRMC for Action Summarization.} Action summarization involves providing a large language model (LLM) with a sequence of actions along with their start and end times. The task is then to use task-oriented prompts to generate the responses. This is a novel task, and there is currently no standard metric for evaluating the effectiveness of action summarization. To address this, we introduce a new metric, Response Meaning Consistency (RMC), to evaluate action summarization performance.

As shown in Fig.~\ref{fig:rmc}, an action sequence~(as) contains both the ground truth actions and their corresponding start and end times. XRFMamba also predicts action categories along with their start and end times. After inputting the sequence into the LLM for question-answering, the LLM generates responses based on both the ground truth and the predictions. When the model's predictions are more accurate, the responses generated for the ground truth and the predictions should be consistent in meaning. To evaluate this, we invite human auditors to determine whether the meaning of the responses is consistent. The RMC can thus be computed as follows:
\begin{equation}\label{eq:rmc}
    RMC = \sum_{i=1}^{N_q} \mathtt{Human}(\mathtt{LLM}(as_{gt}^i, prompt_i), \mathtt{LLM}(as_{pred}^i, prompt_i)) /  {N_q}
\end{equation}
where $\mathtt{Human}$ outputs 1 if the human auditor considers the LLM's responses on ground truth and predicted action sequences share the same meaning. Specifically, for XRF V2, for the 171 test sequences, we prepared two prompts for each sequence, resulting in a total of 342 LLM responses to the ground truth and another 342 responses to the predicted results. All used prompts are presented in the Appendix.~\ref{appsec:prompts}.

We invited five auditors (3 male and 2 female), each with 1-3 years of experience using LLM models, to evaluate whether the meanings of the real and predicted responses were consistent for each of the 342 pairs. As shown in Eq.~\ref{eq:rmc}, when the meanings are consistent, the numerator is incremented by 1, and the denominator represents the total number of evaluations, $N_q=342$. To reduce randomness, we perform action summarization using three different LLM models: ChatGPT-4o, Qwen2.5-Plus, and DeepSeek-V3, under their free plans. The final evaluation metric, mRMC, is obtained by averaging the results from different models and auditors. This provides a more robust and reliable measure for assessing the performance of action summarization. 

For automatic evaluation, we also use LLM models to assess the consistency of each response pair. The prompt we use is \textit{``These are two responses regarding the question of \#\#(Question): (1) \#\#(Response to ground truth) (2) \#\#(Response to the action summary results) . Please determine the consistency between the two answers based on key information (such as whether the action occurs or the number of occurrences of the action), ignoring the difference in the time of the action. If they are consistent, it is 1; otherwise, it is 0''}.  In the prompt, \#\#(Question) are Task-oriented Prompts listed in the Appendix.D.2, such as \textit{ ``Please answer the number of times the user has consumed water during this period?''}; \#\#(Response to ground truth) and \#\#(Response to the action summary results) are the LLM's responses to the ground-truth action sequences and XRFMamba estimated action sequences, respectively.

\begin{table}[ht] 
\caption{Comparison of different methods. We evaluate several state-of-the-art approaches adapted in IMU-based temporal action localization~(TAL)~\cite{bock2024temporal}, Wi-Fi based temporal action detection~\cite{wang2023u} and TAL~\cite{liu2024wificsibasedtemporal}. Our method achieves the best performance, outperforming the recent WiFiTAD~\cite{liu2024wificsibasedtemporal} by a large margin of 5.49 points in mAP@avg, while using 35\% fewer parameters.}
\small
\setlength{\tabcolsep}{2pt}
\begin{tabular}{lccccccccccccc}
\hline
model                & mAP@0.5 & 0.55 & 0.6 & 0.65 & 0.7 & 0.75 & 0.8 & 0.85 & 0.9 & 0.95 & mAP@avg & Params & GFlops \\ \hline
TemporalMaxer~\cite{tang2023temporalmaxer}         & 89.59 & 86.70 & 81.23 & 69.06 & 54.82 & 39.60 & 24.91 & 11.84 & 4.13 & 0.77 & 46.27 & 9.73M &    7.93    \\
ActionFormer~\cite{zhang2022actionformer}         & 90.56 & 87.94 & 83.28 & 77.59 & 68.60 & 54.94 & 37.09 & 18.65 & 5.89 & 0.76 & 52.53 &    47.66M        &   15.70    \\
TriDet~\cite{shi2023tridet}         & 92.90 & 91.34 & 89.51 & 83.37 & 77.91 & 67.50 & 51.46 & 31.19 & 11.61 & 2.28 & 59.91 & 16.12M &    8.94    \\
UWiFiAction~\cite{wang2023u}                 & 87.59    &   86.32    &   84.12     &   81.94     &   78.33     &   66.77    &    53.11    &    34.23    &  13.17      &   1.84     &  58.74  &  6.24M &    -   \\
WiFiTAD~\cite{liu2024wificsibasedtemporal}              & 95.34    &   94.71    &   94.32     &   92.83     &   86.32     &   81.75    &    73.62    &    57.73    &  36.69      &   19.14     &  73.25  &  121.06M          &    44.81    \\
ActionMamba~\cite{chen2024video-mamba-suite}         & 95.05 & 94.84 & 94.10 & 92.81 & 91.54 & 84.63 & 76.56 & 57.54 & 26.89 & 4.35 & 71.83 &  29.35M          &   11.91     \\
XRFMamba~(ours) &  \textbf{96.67} &   \textbf{96.23}      &    \textbf{95.68}    &    \textbf{94.53}     &   \textbf{92.02}     &     \textbf{88.66}    &    \textbf{77.92}    &     \textbf{63.96}    &  \textbf{46.53}      &   \textbf{35.13}      &  \textbf{78.74}  &    78.38M        &   43.49  \\ \hline  
\end{tabular}
\label{tab:compare-with-other-methods}
\end{table}

\subsection{Results}\label{sec:results}

\begin{figure}[ht]
    \centering
    \begin{minipage}{0.63\textwidth}
        \centering
        \includegraphics[width=\linewidth]{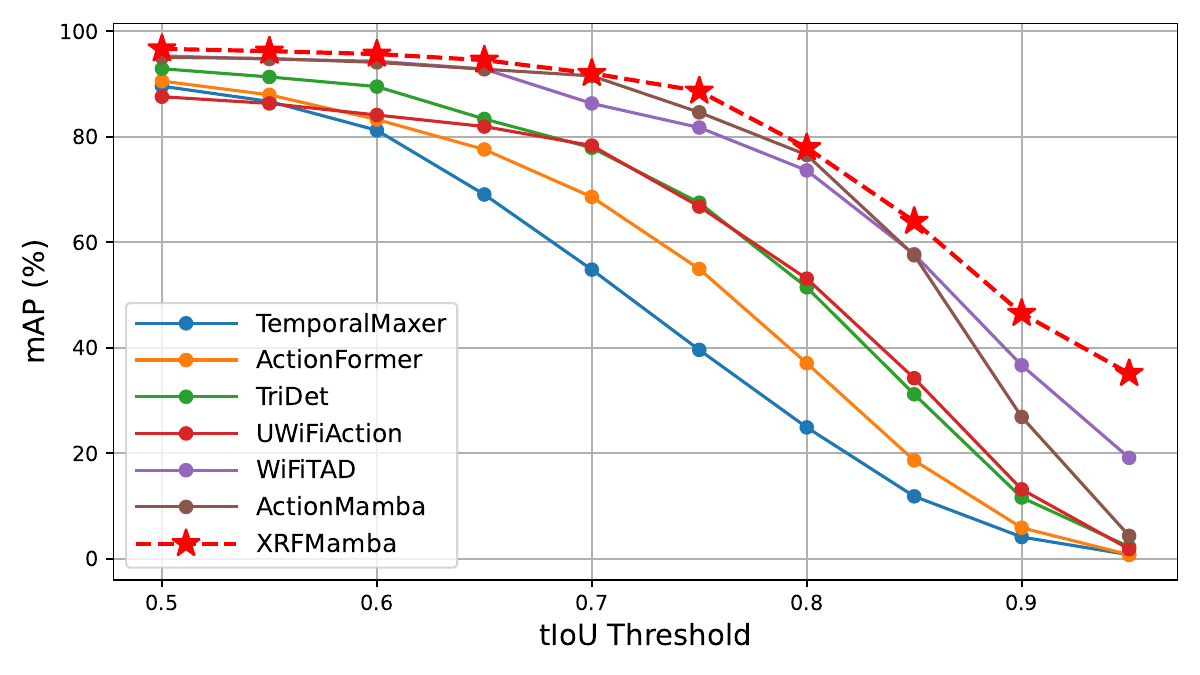}
        \caption*{(a) mAP comparison across tIoU thresholds}
    \end{minipage}
    \hspace{-0.25cm}
    \begin{minipage}{0.35\textwidth}
        \centering
        \includegraphics[width=\linewidth]{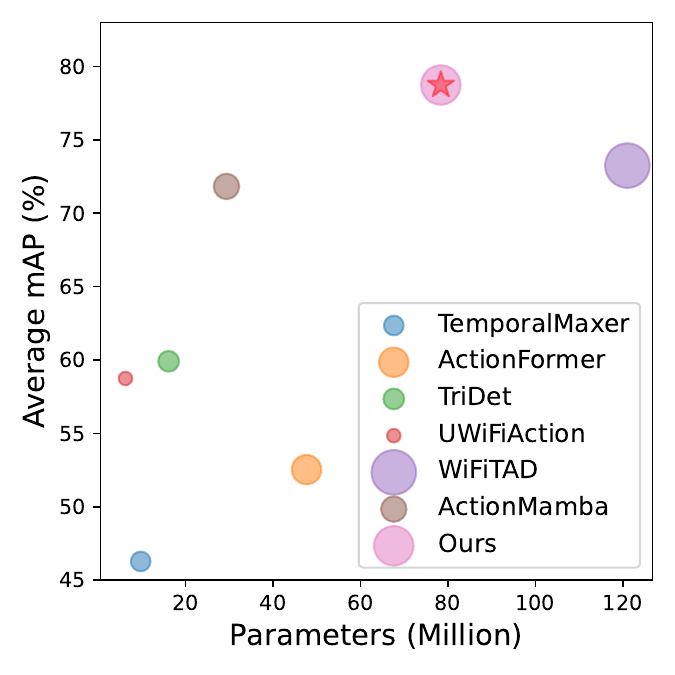}
        \caption*{(b) Accuracy-efficiency trade-off}
    \end{minipage}
    \caption{Visual comparison of XRFMamba and existing approaches.  
    (a) presents a comparison of mAP across different tIoU thresholds, showing that XRFMamba consistently outperforms state-of-the-art methods.  
    (b) illustrates the trade-off between precision, model parameters, and computational complexity (bubble radius, FLOPs), highlighting the balance achieved by XRFMamba in performance and efficiency.}
    \label{fig:method-performance-r4}
\end{figure}

\begin{figure}[t]
    \centering
    \begin{subfigure}{\linewidth}
        \centering
        \includegraphics[width=0.95\linewidth]{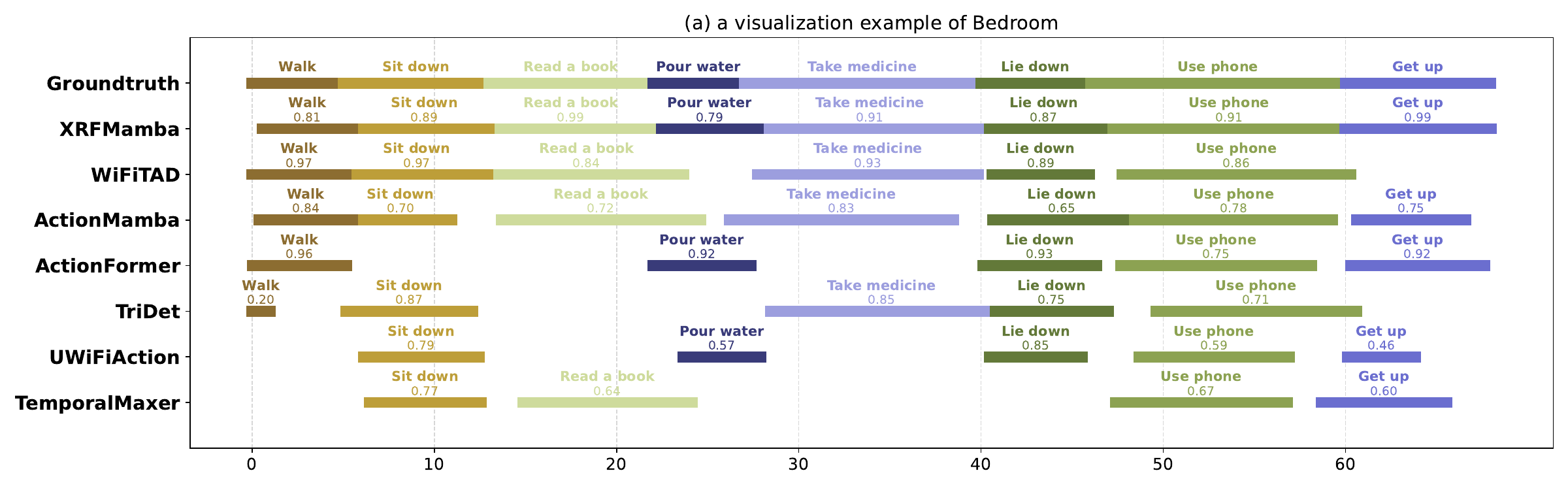}
        \label{fig:sub1}
    \end{subfigure}
    \vspace{-0.2cm} 
    \begin{subfigure}{\linewidth}
        \centering
        \includegraphics[width=0.95\linewidth]{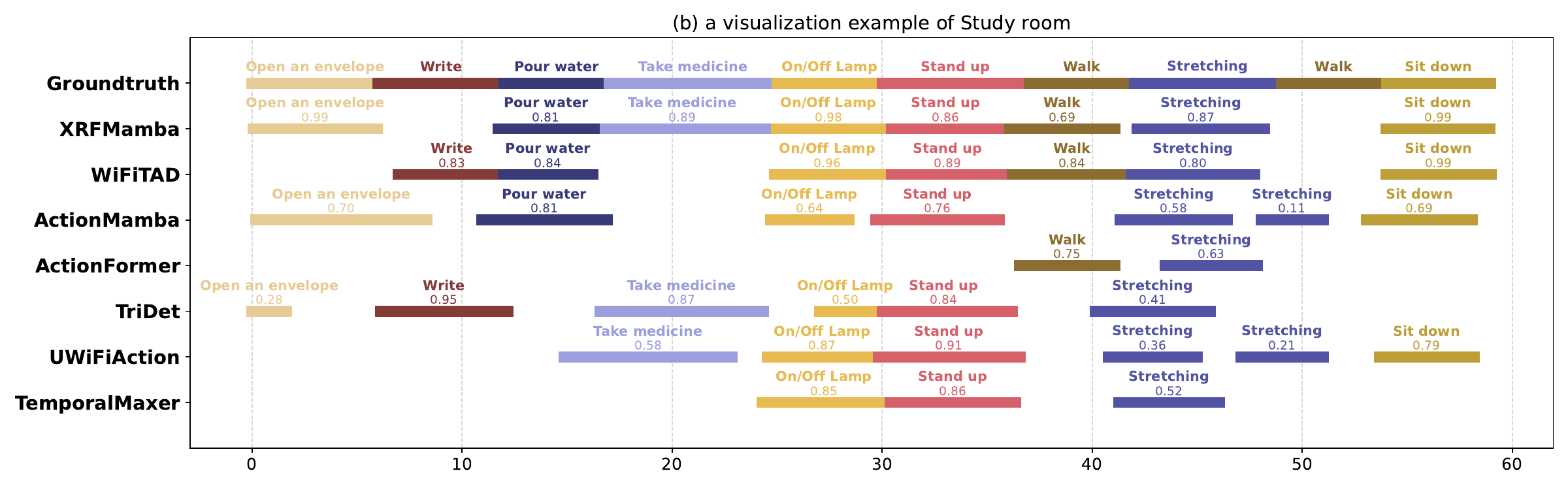}
        \label{fig:sub2}
    \end{subfigure}
    \vspace{-0.2cm} 
    \begin{subfigure}{\linewidth}
        \centering
        \includegraphics[width=0.95\linewidth]{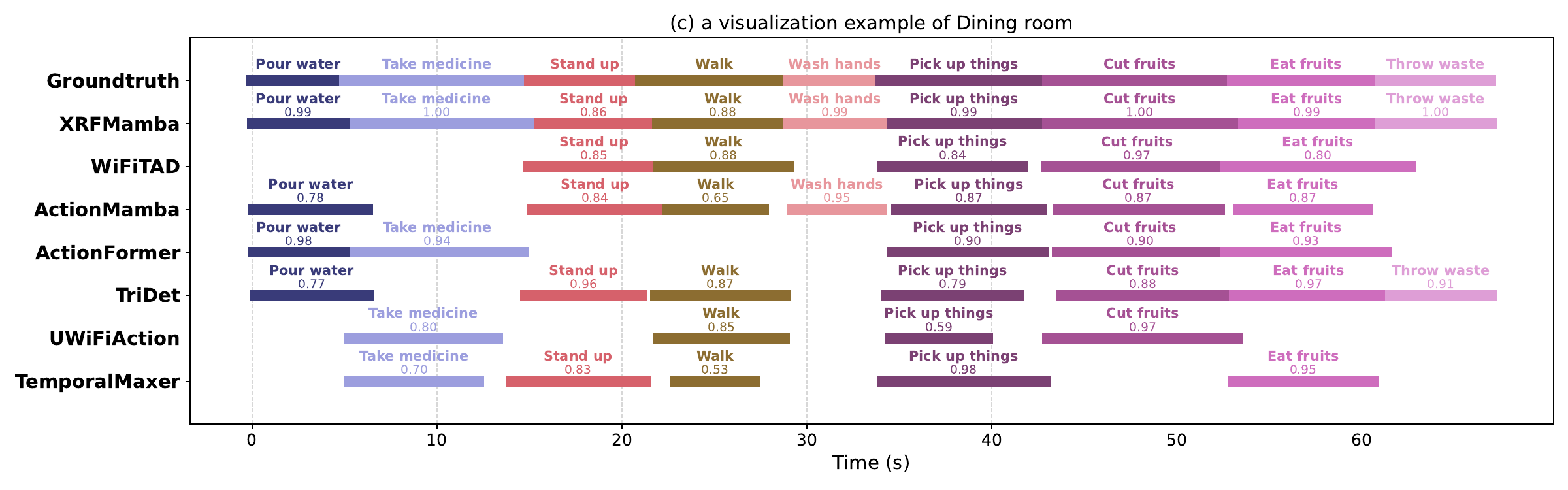}
        \label{fig:sub3}
    \end{subfigure}
    \vspace{-0.5cm} 
    \caption{Example of the predicted results in Bedroom, Study room, and Dining room. In all the scenes, XRFMamba demonstrates excellent performance, with superior classification accuracy and precise action boundary localization compared to state-of-the-art methods.}
    \label{fig:visualization}
\end{figure}

\subsubsection{Temporal Action Localization Results}\label{sec:tal-results}

We evaluate several state-of-the-art approaches, including TriDet~\cite{shi2023tridet}, ActionFormer~\cite{zhang2022actionformer}, and TemporalMaxer~\cite{tang2023temporalmaxer}, which have been recently applied to IMU-based temporal action localization~\cite{bock2024temporal}. We also compare with UWiFiAction~\cite{wang2023u}, a WiFi-based action detection method, and WiFiTAD~\cite{liu2024wificsibasedtemporal}, a WiFi-based temporal action localization method. Additionally, we test ActionMamba~\cite{chen2024video-mamba-suite}, an advanced Mamba-based model for video-based action localization. 

To ensure a fair comparison, all methods are extended to support dual-input (IMU and WiFi) fusion. Specifically, as in the comparison experiment with WiFiTAD~\cite{liu2024wificsibasedtemporal}, all methods use the same inputs, embedding, TSSE, and prediction head as  XRFMamba, with the only difference being that XRFMamba's backbone is replaced by the backbone of these methods.
   As shown in Table~\ref{tab:compare-with-other-methods}, our method achieves the best performance, outperforming the recently accepted WiFiTAD~\cite{liu2024wificsibasedtemporal} by a large margin of 5.49 points in mAP@avg, while using 35\% fewer parameters. We also conducted a t-test evaluation, which demonstrates that XRFMamba significantly outperforms WiFiTAD, with detailed analysis provided in  Appendix~\ref{app: Significance Analysis}. Furthermore, we benchmark the runtime performance of XRFMamba on an NVIDIA RTX 3090 GPU. XRFMamba achieves an average inference latency of $37.43 \pm 34.61$ ms per 30-second window, significantly lower than WiFiTAD’s $58.63 \pm 32.80$ ms. Additionally, XRFMamba uses less peak GPU memory (660.78 MB vs. 827.26 MB), demonstrating superior efficiency in both processing time and memory usage. XRFMamba also shows improved efficiency in processing entire sequences, with an average sequence time of $536.75 \pm 167.39$ ms, compared to WiFiTAD’s $840.67 \pm 214.53$ ms. This allows the system to process an entire sequence within one second, which is crucial for real-time applications. These results highlight the effectiveness of our method and the overall efficiency of XRFMamba, making it well-suited for deployment on resource-constrained platforms.

Fig.~\ref{fig:method-performance-r4}(a) presents a line chart showing the mAP performance of our method and baseline approaches under various tIoU thresholds. This provides a more intuitive understanding of how our method consistently outperforms others across different levels of localization strictness. Fig.~\ref{fig:method-performance-r4}(b) provides a bubble chart comparing our method with prior works in terms of accuracy (average mAP), model complexity (number of parameters), and computational cost (FLOPs). The x-axis represents model size, the y-axis represents performance, and the bubble radius corresponds to computational cost, helping to reveal the trade-off between accuracy and efficiency.

Fig.~\ref{fig:visualization} presents prediction examples in Bedroom, Study room, and Dining room, with comparison methods including WiFiTAD~\cite{liu2024wificsibasedtemporal}, ActionMamba~\cite{chen2024video-mamba-suite}, ActionFormer~\cite{zhang2022actionformer}, TriDet~\cite{shi2023tridet}, UWiFiAction~\cite{wang2023u}, and TemporalMaxer~\cite{tang2023temporalmaxer} from Table~\ref{tab:compare-with-other-methods}. It is visually clear that WiFiTAD, ActionMamba, and our XRFMamba perform better, with XRFMamba demonstrating superior classification accuracy and precise action boundary localization compared to state-of-the-art methods. In terms of practical applications, more accurate action classification and boundary localization enable the system to respond more precisely to user actions, along with a more accurate action summary. This results in better service, fewer false positives, and significantly enhances the performance and functionality of the system, making it more intelligent, reliable, and user-friendly.

We further present the mAP (mean Average Precision) for each action in Table~\ref{tab:map-per-action}. From the table, it is clear that the mAP@avg for most actions remains relatively high, indicating that XRFMamba performs well in recognizing and localizing the majority of actions. However, some actions, such as Stretching and Opening an Envelope, have relatively low mAP values compared to others. The Stretching action involves various body parts and features a wide range of motion with substantial variation in execution style. Different individuals may perform the stretching action in vastly different postures, which increases the complexity of recognizing the action accurately. In contrast, Opening an Envelope typically consists of subtle, short-duration hand movements, often involving just the fingers and palms. These localized and brief movements are more challenging to capture. Additionally, there is a sudden drop at 0.75 and 0.8 for Lying Still because it appeared only once in the test data, with its tIoUS being 0.7564.

\begin{table}[ht]
\small
 \caption{mAP of 30 actions. mAP@avg for most actions remains relatively high indicating that XRFMamba performs well in XRF V2 dataset for temporal action localization. }
 \setlength{\tabcolsep}{3.2pt}
\begin{tabular}{lccccccccccc}
\hline
action & mAP@0.5 & 0.55 & 0.6 & 0.65 & 0.7 & 0.75 & 0.8 & 0.85 & 0.9 & 0.95 & mAP@avg \\ \hline
Walk & 99.65 & 99.33 & 99.33 & 99.33 & 95.20 & 91.55 & 87.82 & 72.56 & 52.50 & 45.95 & 84.32 \\
Sit down & 99.34 & 98.35 & 98.13 & 98.13 & 97.26 & 97.26 & 92.57 & 73.12 & 63.52 & 56.61 &  87.43 \\
Stand up & 99.88 & 99.88 & 99.88 & 99.88 & 94.93 & 91.83 & 81.24 & 68.52 & 48.68 & 41.90 & 82.66 \\
Pour water into the cup & 99.75 & 98.72 & 98.72 & 98.52 & 95.48 & 91.85 & 88.08 & 69.84 & 47.34 & 43.83 & 83.21 \\
Drink water & 94.99 & 94.99 & 94.99 & 94.99 & 88.97 & 88.97 & 88.64 & 69.76 & 56.24 & 38.10 & 81.07 \\
Take medicine & 98.17 & 98.17 & 98.17 & 98.17 & 96.67 & 92.60 & 85.35 & 75.52 & 62.11 & 46.03 & 85.10 \\
Pick up things & 98.73 & 98.73 & 98.73 & 98.73 & 98.73 & 98.73 & 85.18 & 67.18 & 58.82 & 57.18 & 86.07 \\
Take the fruits from the cabinet & 100.00 & 100.00 & 100.00 & 100.00 & 88.74 & 72.96 & 72.14 & 57.89 & 36.78 & 33.25 & 76.18 \\
Cut fruits & 99.75 & 99.75 & 99.75 & 94.87 & 94.29 & 89.80 & 82.64 & 55.46 & 24.33 & 15.75 & 75.64 \\
Eat fruits & 98.45 & 98.45 & 96.10 & 94.48 & 94.06 & 91.38 & 79.61 & 66.79 & 45.47 & 26.76 & 79.16 \\
Wash hands & 99.74 & 99.74 & 99.74 & 98.15 & 97.71 & 95.71 & 93.72 & 75.37 & 53.69 & 46.99 & 86.06 \\
Throw waste & 97.79 & 97.79 & 97.79 & 97.79 & 95.25 & 95.25 & 89.08 & 84.92 & 74.85 & 71.16 &  90.16 \\
Wipe the table & 97.55 & 94.58 & 92.79 & 89.40 & 89.30 & 89.30 & 83.68 & 83.68 & 72.55 & 59.04 & 85.19 \\
Stretching & 53.93 & 53.21 & 52.59 & 46.07 & 42.64 & 37.03 & 27.53 & 21.26 & 7.90 & 1.87 & \uwave{34.40} \\
Turn on and off the desk lamp  & 98.42 & 98.42 & 98.42 & 97.46 & 95.54 & 95.54 & 85.38 & 69.30 & 60.77 & 54.82 & 85.41 \\
Operate the mouse & 96.43 & 96.43 & 96.43 & 96.43 & 96.43 & 87.42 & 83.49 & 65.83 & 59.69 & 39.77 & 81.84 \\
Write & 99.62 & 99.62 & 99.62 & 95.18 & 90.23 & 87.38 & 86.10 & 64.30 & 46.22 & 38.06 & 80.63 \\
Operate the keyboard & 95.97 & 95.97 & 95.97 & 95.97 & 95.97 & 90.98 & 85.58 & 76.38 & 60.35 & 53.76 & 84.69 \\
Read a book & 97.39 & 97.39 & 97.39 & 97.34 & 94.79 & 94.79 & 94.16 & 85.60 & 47.73 & 23.79 & 83.04 \\
Open an envelope & 83.14 & 75.93 & 75.07 & 70.76 & 64.49 & 59.00 & 55.48 & 43.17 & 16.88 & 12.14 & \uwave{55.61} \\
Answer the phone & 100.00 & 100.00 & 100.00 & 100.00 & 100.00 & 95.26 & 86.28 & 54.70 & 23.06 & 16.79 & 77.61 \\
Write on the blackboard & 99.43 & 99.43 & 99.43 & 99.43 & 96.40 & 94.29 & 94.29 & 88.82 & 82.64 & 50.75 &  90.49 \\
Get up & 95.42 & 95.42 & 95.42 & 95.42 & 95.42 & 89.25 & 89.25 & 81.90 & 52.41 & 28.52 & 81.84 \\
Lie down & 99.71 & 99.71 & 99.71 & 93.12 & 87.32 & 80.84 & 55.82 & 48.48 & 34.96 & 15.35 & 71.50 \\
Use phone & 99.11 & 99.11 & 99.11 & 99.11 & 99.11 & 91.91 & 85.12 & 35.15 & 19.80 & 5.43 & 73.29 \\
Open and close windows & 99.77 & 99.77 & 99.77 & 99.77 & 90.75 & 89.51 & 58.42 & 54.33 & 45.90 & 38.86 & 77.69 \\
Open and close curtains & 100.00 & 100.00 & 100.00 & 100.00 & 100.00 & 96.34 & 81.87 & 71.25 & 53.17 & 41.35 & 84.40 \\
Water plants & 99.74 & 99.74 & 99.74 & 99.74 & 97.35 & 97.35 & 88.58 & 69.34 & 51.35 & 26.03 & 82.89 \\
Stand still & 98.21 & 98.21 & 87.71 & 87.71 & 87.71 & 85.71 & 59.49 & 57.31 & 36.20 & 24.17 & 72.25 \\
Lying still  & 100.00 & 100.00 & 100.00 & 100.00 & 100.00 & 100.00 & 0.00 & 0.00 & 0.00 & 0.00 & \uwave{62.22} \\
$\textbf{mean}$ & 96.67 & 96.23 & 95.68 & 94.53 & 92.02 & 88.66 & 77.55 & 63.59 & 46.53 & 35.13 & 78.73 \\\hline
\end{tabular}
\label{tab:map-per-action}
\end{table}

The primary limitation of our method lies in the number of parameters that need to be learned, which is nearly 2.67 times that of ActionMamba~\cite{chen2024video-mamba-suite}. This is largely due to the use of TSSE (Temporal Signal Semantic Embedding) embedding module from WiFiTAD~\cite{liu2024wificsibasedtemporal}, which accounts for approximately 55.5M parameters. Therefore, a promising future research direction is to explore lighter and more efficient embedding modules that can maintain performance while significantly reducing parameters. Additionally, considering that the current state-of-the-art in video-based temporal action localization (TAL) method, such as ActionMamba~\cite{chen2024video-mamba-suite}, often relies on pre-trained models, such as InterVideo-6B~\cite{Internvideo}, as embedding modules, another potential research direction is to investigate the use of pre-trained time-series foundation models, such as TimesFM~\cite{TimesFM}, FoundTS~\cite{Foundts}, and Time-MoE~\cite{TimeMOE}, as the embedding module, thereby reducing the number of parameters that need to be learned during training.

\subsubsection{Action Summarization Results}\label{sec:action-summarization-results}

\begin{table}[ht] 
 \caption{XRFMamba achieves high  response meaning consistency (RMC) in action summarization. \textit{(R3 and R4)}
  }
\begin{tabular}{lcccc}
\hline
      & ChatGPT-4o  & Qwen2.5-Plus & DeepSeek-V3 &  mean \\ \hline
RMC@a1 & [278/342]0.813 & [276/342]0.807 & [278/342]0.813 & 0.811 \\
RMC@a2 & [276/342]0.807 & [275/342]0.804 & [276/342]0.807 & 0.806 \\
RMC@a3 & [279/342]0.816 & [279/342]0.816 & [276/342]0.807 & 0.813 \\
RMC@a4 & [278/342]0.813 & [275/342]0.804 & [277/342]0.810 & 0.809  \\
RMC@a5 & [262/342]0.766 & [263/342]0.769 & [266/342]0.778 & 0.771 \\
mRMC(humans)  & 0.803  & 0.800 & 0.803 & 0.802 \\ \hline
RMC@ChatGPT-4o & [271/342]0.792 & [269/342]0.787 & [275/342]0.804 & 0.794\\
RMC@Qwen2.5-Plus & [273/342]0.798 & [265/342]0.775 & [280/342]0.819 & 0.797 \\
RMC@DeepSeek-V3 & [271/342]0.792 & [265/342]0.775 & [276/342]0.807 & 0.791 \\
mRMC(LLMs)  & 0.794 & 0.779 & 0.810 & 0.794 \\ \hline
\end{tabular}
\label{tab:rmc-results}
\end{table}

\begin{figure}[ht]
    \centering
    \includegraphics[width=0.8\linewidth]{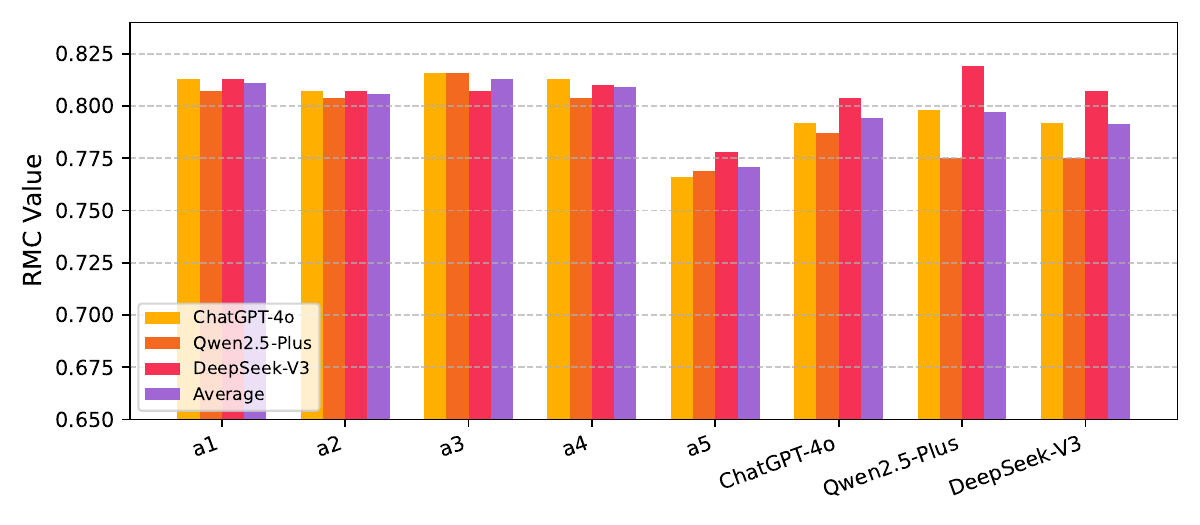}
    \caption{RMC scores of three LLM agents across auditor groups. XRFMamba shows consistently high performance around 0.8.}
    \label{fig:visualization-table6-r4}
\end{figure}

We perform action summarization using three different LLM models: ChatGPT-4o, Qwen2.5-Plus, and DeepSeek-V3, all under their free plans. The consistency between predicted and real action summaries from five auditors is presented in Table~\ref{tab:rmc-results} and Fig~\ref{fig:visualization-table6-r4}. As shown in the table, auditors 1 to 4 consistently acknowledge that our LLMs' predictions for TAL demonstrate high alignment with ground truth, achieving a response meaning consistency~(RMC) of approximately 0.81. Auditor 5 exhibits slightly lower agreement with an RMC of 0.77. The mRMC across all human auditors reaches 0.802, indicating that XRFMamba achieves strong performance in both temporal action localization and action summarization. The overall mRMC across all LLM-based evaluators is 0.794, which is closely aligned with the human assessment, demonstrating that the automated evaluation effectively replicates human judgments.

\begin{figure}[ht]
    \centering
    \includegraphics[width=1\linewidth]{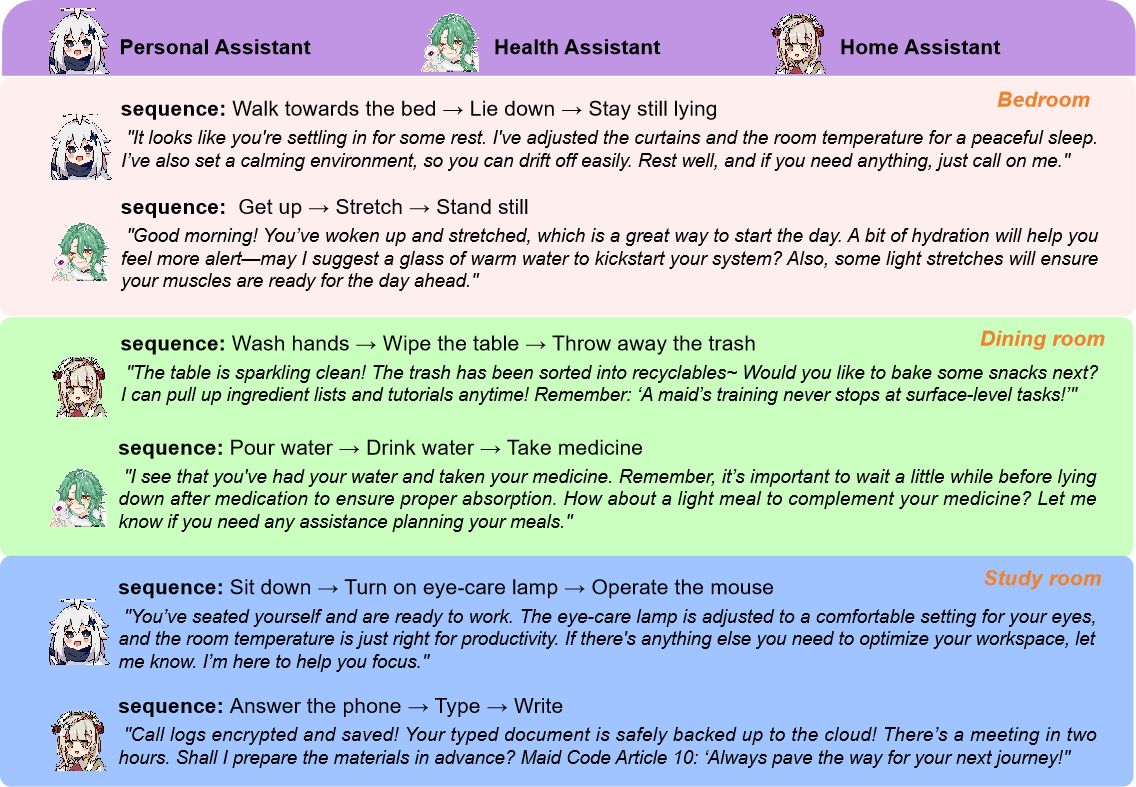}
    \caption{XRFMamba, when combined with LLMs, can function as an intelligent ambient sensing agent, such as a personal assistant, health assistant, and home assistant,  to provide services like task reminders, personalized health monitoring, temperature control, and more.}
    \label{fig:smart-agent}
\end{figure}

Notably, the auditor-specific RMC values show minimal correlation with LLM selection, indicating that XRFMamba, when integrated with free-plan LLMs, can effectively function as an intelligent ambient sensing agent. Practical implementations of such agents, including personal assistants, health monitoring systems, and smart home controllers (as illustrated in Fig.~\ref{fig:smart-agent}), demonstrate capabilities ranging from task management and personalized health tracking to indoor temperature control, thereby significantly enhancing smart home ecosystems through multifunctional intelligent services.

\subsubsection{Leave-one-person-out Results}\label{sec:leave-one-person-out-results}

In real-world applications, XRFMamba should not be limited to specific individuals but should instead be adaptable to a wide range of users with varying characteristics, such as body movement patterns, walking styles, and device usage habits. To assess this adaptability to new users, we employ a leave-one-person-out cross-validation evaluation. In this evaluation, data from one volunteer is used as the test set, while data from all other users are used for training. As shown in Table~\ref{tab:leave-one-person-out-results}, mAP for new users is 72.84, indicating that our method performs well and is capable of generalizing effectively to unseen users. 

\begin{table}[t]
\centering
\caption{Leave-one-person-out results. mAP for new users reaches 72.84, indicating that XRFMamba performs well and is capable of generalizing effectively to unseen users.}
\begin{tabular}{lccccccccccc}
\hline
Test user & mAP@0.5 & 0.55 & 0.6 & 0.65 & 0.7 & 0.75 & 0.8 & 0.85 & 0.9 & 0.95 & mAP@avg \\ \hline
user1 & 94.31 & 94.29 & 93.96 & 93.91 & 93.04 & 90.85 & 84.06 & 74.31 & 53.53 & 43.54 & 81.58 \\
user2 & 90.08 & 88.89 & 87.84 & 84.72 & 80.64 & 75.15 & 61.94 & 44.69 & 27.68 & 20.13 & 66.18 \\
user3 & 96.91 & 96.91 & 96.49 & 96.14 & 92.21 & 88.25 & 84.93 & 75.87 & 57.34 & 45.85 & 83.09 \\
user4 & 96.97 & 96.61 & 95.57 & 94.62 & 90.71 & 85.03 & 76.84 & 61.34 & 41.90 & 25.27 & 76.49 \\
user5 & 55.42 & 55.30 & 54.79 & 53.31 & 51.18 & 48.45 & 44.80 & 36.64 & 23.37 & 14.67 & 43.79 \\
user6 & 92.63 & 92.16 & 90.86 & 89.57 & 87.91 & 86.37 & 83.59 & 74.51 & 61.00 & 45.90 & 80.45 \\
user7 & 91.15 & 89.98 & 89.12 & 87.78 & 85.84 & 82.33 & 75.20 & 62.78 & 47.43 & 39.66 & 75.13 \\
user8 & 90.76 & 90.64 & 90.07 & 89.41 & 84.24 & 78.33 & 71.25 & 58.67 & 39.68 & 31.31 & 72.44 \\
user9 & 93.44 & 93.12 & 91.71 & 90.70 & 87.47 & 83.60 & 76.70 & 60.40 & 41.77 & 32.18 & 75.11 \\
user10 & 89.21 & 88.21 & 87.14 & 84.68 & 77.40 & 72.43 & 61.59 & 45.25 & 31.87 & 24.79 & 66.26 \\
user11 & 95.71 & 95.38 & 94.69 & 93.30 & 89.27 & 85.40 & 79.47 & 65.21 & 44.65 & 35.75 & 77.88 \\
user12 & 92.55 & 92.09 & 90.98 & 89.84 & 85.21 & 78.49 & 71.52 & 56.59 & 35.82 & 25.18 & 71.83 \\
user13 & 95.45 & 95.45 & 94.81 & 94.20 & 91.45 & 87.77 & 81.00 & 68.42 & 47.25 & 34.38 & 79.02 \\
user14 & 91.23 & 91.22 & 90.65 & 88.31 & 84.83 & 81.50 & 74.48 & 62.79 & 46.23 & 36.23 & 74.75 \\
user15 & 86.24 & 85.77 & 85.07 & 82.19 & 77.43 & 68.54 & 64.21 & 49.70 & 36.31 & 27.63 & 66.31 \\
user16 & 93.38 & 92.69 & 91.00 & 88.75 & 85.44 & 81.84 & 73.44 & 60.77 & 47.54 & 35.82 & 75.07 \\
Average & 90.34 & 89.92 & 89.05 & 87.59 & 84.02 & 79.65 & 72.81 & 59.87 & 42.71 & 32.39 & 72.84 \\
\hline
\end{tabular}
\label{tab:leave-one-person-out-results}
\end{table}

\begin{figure}[ht]
    \centering
    \includegraphics[width=0.9\linewidth]{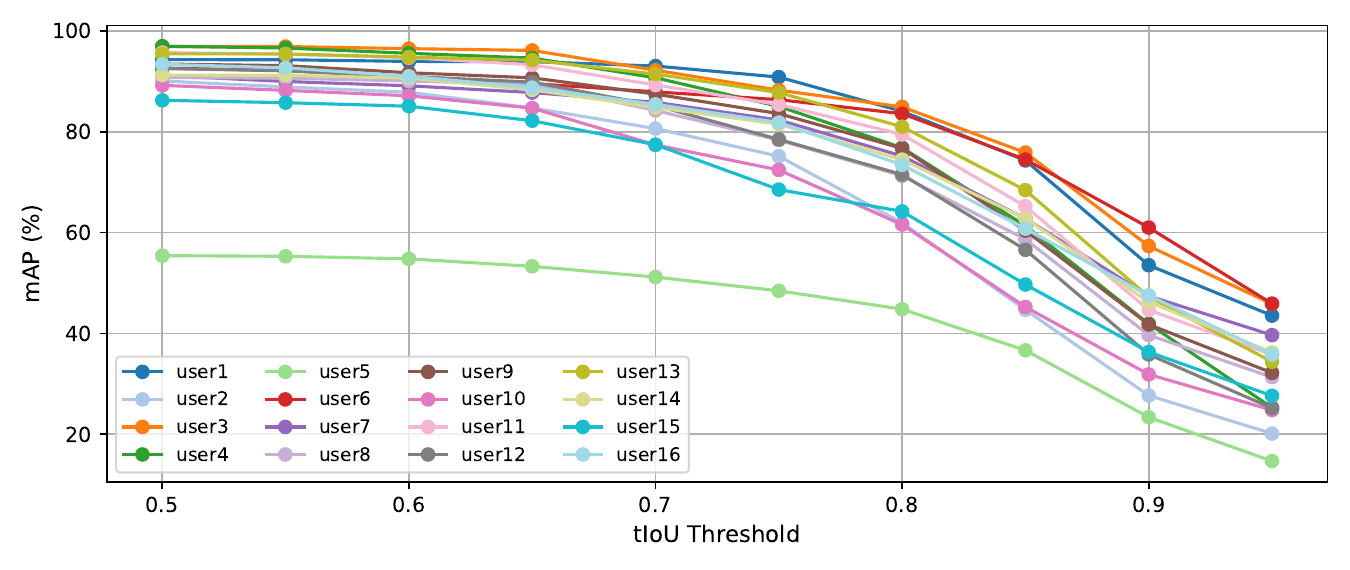}
    \caption{Performance of XRFMamba in a leave-one-person-out evaluation.}
    \label{fig:leave-one-person-out-results} 
\end{figure}

Fig.~\ref{fig:leave-one-person-out-results} illustrates the performance of XRFMamba under a leave-one-person-out cross-validation scheme. The overall stability of the curve indicates that our method consistently maintains high recognition accuracy, even when encountering new users. However, we observed that the performance of User 5 was noticeably worse compared to the other users. Upon reviewing the synchronized video, we discovered that User 5 exhibited certain behavior patterns and usage habits that differed significantly from those of the other participants. For example, User 5 tended to open windows with their left hand, while the others used their right hand. These discrepancies, including variations in walking speed, body movements, or device handling, likely contributed to the poorer performance observed for this user in the leave-one-person-out evaluation. Despite this outlier, the model demonstrated strong generalization capability, as evidenced by the overall stability and consistency of the performance for the remaining users.

In our experiments, we have not performed cross-environment testing because the actions differ across environments, as shown in Table~\ref{tab:proposed-action}. Actions in one environment may not exist in another, making cross-environment testing inconsistent and unreliable. Therefore, it is not feasible in our current setup.

\subsection{Ablation Study}\label{sec:ablation-study}

\subsubsection{How and Where to Conduct Fusion}\label{sec:fusion}

XRFMamba takes both IMU and Wi-Fi modalities from XRF V2 as input. We first test where and how modality fusion should be applied in the XRFMamba pipeline. We evaluate fusion after the projection, after embedding, and after the Mamba backbone. In terms of fusion strategies, we evaluate three methods: (1) simple weighted addition: applying fixed weight coefficients to the IMU (0.8) and Wi-Fi (0.2) features and directly adding the weighted features;  (2) linear fusion: combining the IMU and Wi-Fi features through a learned linear transformation, followed by a gating mechanism using a sigmoid function to weight the contribution of each modality;  and (3) gate fusion: applying separated convolutional layers to the IMU and Wi-Fi features to compute individual gating weights using a sigmoid function, which are then used to weight each modality and combine them through summation. For a detailed explanation of each fusion method, including the mathematical formulations and pseudocode, please refer to the Appendix.~\ref{appsec:fusion-codes}.

The results of the nine different combinations of fusion positions and strategies are shown in Table~\ref{tab:fusion-ablation}. As illustrated in the table, fusion applied later in the pipeline involves more separate data streams and additional parameters. When comparing different fusion strategies, the gate and linear methods require more parameters than the weighted method, yet their performance is not as effective. In terms of fusion positions, combination 1, which applies weighted fusion after the projection module, achieves an average mAP of 78.74. In contrast, combination 3, which performs fusion after the Mamba backbone, achieves a higher mAP of 80.89 but requires nearly double the number of parameters. Therefore, we chose the weighted addition fusion method applied after the projection module as the final fusion approach for XRFMamba, as it offers the best mAP-parameter trade-off. This trade-off is particularly important for ensuring the model's suitability to deliver fast, reliable, and resource-efficient operation in deployment environments.

\begin{table}[ht]
\small
\caption{We selected the weighted addition fusion method applied after the projection module (\#1) as the final fusion approach for XRFMamba, as it achieves the best mAP-parameter trade-off.}
\setlength{\tabcolsep}{2.5pt}
\begin{tabular}{c llccccccccccccc}
\hline
\# & fusion & fusion after & mAP@0.5 & 0.55 & 0.6 & 0.65 & 0.7 & 0.75 & 0.8 & 0.85 & 0.9 & 0.95 & mAP@avg & Params & Gflops\\ \hline
1 & weight & projection      &  96.67 &   96.23      &    95.68    &    94.53     &   92.02     &     88.66    &    77.92    &     63.96    &  46.53      &   35.13      &  \uwave{78.74}  &   \uwave{78.38M}  & \uwave{43.49} \\
2 & weight & embedding     &  96.97 &   96.53      &    95.88    &    95.04     &   92.66     &     88.08    &    80.26    &     67.97    &  49.71      &   38.51      &  80.16  &  128.89M  & 77.58 \\
3 & weight & backbone &  97.12 &   96.83      &    96.57    &    95.88     &   92.91     &     89.80    &    82.69    &     69.48    &  49.42      &   38.19      &  \underline{80.89}  &  \underline{150.08M}  & \underline{81.96} \\
4 & linear & projection      &  94.87 &   94.04      &    93.34    &    91.83     &   87.63     &     83.03    &    76.53    &     62.03    &  40.87      &   34.26      &  75.84  &  95.16M & 52.09 \\
5 & linear & embedding     &  96.23 &   95.59      &    95.04    &    94.41     &   90.68     &     86.99    &    80.17    &     63.65    &  42.50      &   30.45      &  77.57  &  129.14M & 77.71 \\
6 & linear & backbone &  95.86 &   95.50      &    94.94    &    90.68     &   89.13     &     83.60    &    75.69    &     59.11    &  34.07      &   08.08      &  72.66  &  150.43M & 82.14 \\
7 & gate   & projection      &  96.86 &   96.56      &    95.78    &    91.99     &   89.69     &     86.42    &    79.47    &     65.07    &  45.58      &   36.21      &  {78.36}  &  {79.17M} & {44.57} \\
8 & gate   & embedding     &  97.22 &   96.87      &    96.81    &    95.88     &   89.88     &     84.98    &    80.19    &     66.09    &  48.58      &   38.72      &  79.52  &  129.67M  & 77.85 \\
9 & gate   & backbone &  96.92 &   96.69      &    96.41    &    95.58     &   93.27     &     86.03    &    81.01    &     67.56    &  52.83      &   40.93      &  80.72  &  154.03M & 82.22 \\ \hline
\end{tabular}
 \label{tab:fusion-ablation}
\end{table}

\begin{figure}[t]
    \centering
    \begin{minipage}{0.7\textwidth} 
        \centering
        \includegraphics[width=\linewidth]{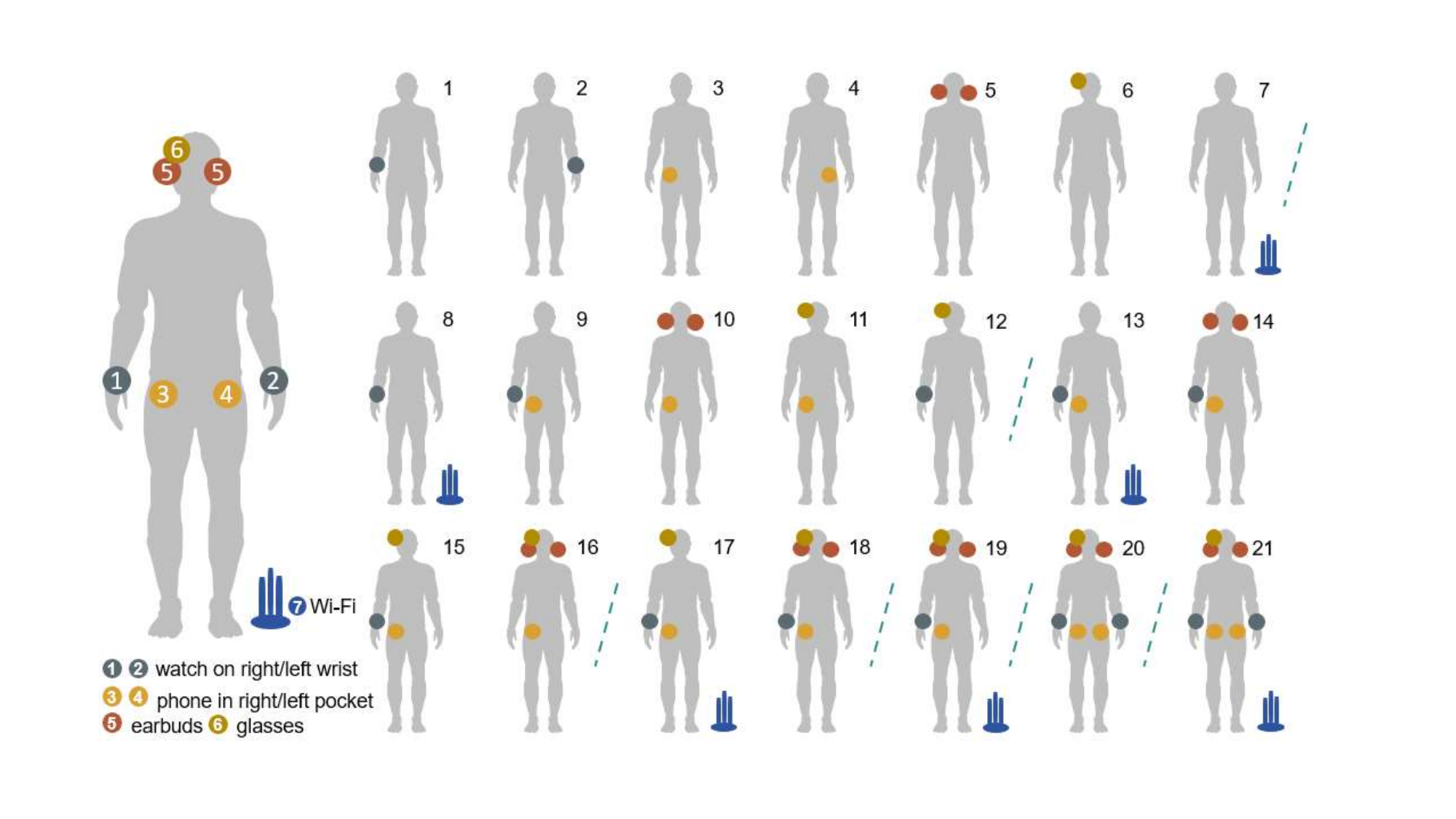}
    \end{minipage}
    \vspace{-0.3cm}
    \begin{minipage}{1\textwidth}
        \centering
        \begin{tabular}{lccccccccccc}
        \hline
        device & mAP@0.5 & 0.55 & 0.6 & 0.65 & 0.7 & 0.75 & 0.8 & 0.85 & 0.9 & 0.95 & mAP@avg \\ \hline
        1      & 95.88 & 95.32 & 93.52 & 90.90 & 85.02 & 76.06 & 60.50 & 40.57 & 17.40 & 2.93 & 65.81 \\
        2      & 94.07 & 93.03 & 91.17 & 87.33 & 79.99 & 67.60 & 51.39 & 27.82 & 10.97 & 1.75 & 60.51 \\
        3      & 80.99   & 79.94 & 77.65 & 74.81 & 71.11 & 64.19 & 53.78 & 37.90 & 19.02 & 2.86 & 56.22 \\
        4      & 91.91 & 90.38 & 85.54 & 79.48 & 68.81 & 54.26 & 35.51 & 18.48 & 7.98 & 1.16 & 53.35 \\
        5  & 85.76          & 83.82      & 80.99      & 76.34       & 70.48      & 56.36       & 43.89      & 29.39       & 13.03      & 2.76        & 54.28         \\
        6  & 89.06          & 88.14      & 86.48      & 83.64       & 75.66      & 66.33       & 50.27      & 30.88       & 9.77       & 1.42        & 58.16         \\
        7  & 65.40          & 60.51      & 54.67      & 47.75       & 34.70      & 24.68       & 13.17      & 5.52        & 1.67       & 0.32        & 30.84         \\
        8      & 95.58 & 95.07 & 93.77 & 88.44 & 85.46 & 77.62 & 67.22 & 49.08 & 26.28 & 4.91 & 68.34 \\
        9      & 95.19   & 94.60 & 92.87 & 90.86 & 86.38 & 78.39 & 65.48 & 46.95 & 18.51 & 4.58 & 67.38 \\
        10     & 91.03   & 89.70 & 87.84 & 82.18 & 78.23 & 71.02 & 57.76 & 37.97 & 18.55 & 4.59 & 61.89 \\
        11     & 92.40   & 91.80 & 90.77 & 87.64 & 84.49 & 76.19 & 65.69 & 49.87 & 28.32 & 7.66 & 67.48 \\
        12     & 94.86 & 94.02 & 93.50 & 91.69 & 89.33 & 82.84 & 71.61 & 52.37 & 22.11 & 3.06 & 69.54 \\
        13     & 95.29 & 94.60 & 93.57 & 92.19 & 89.82 & 83.28 & 68.06 & 44.47 & 18.34 & 2.76 & 68.24 \\
        14     & 94.36 & 94.15 & 92.62 & 90.66 & 87.59 & 80.33 & 68.98 & 51.75 & 31.19 & 10.64 & 70.23 \\
        15     & 93.05   & 92.76 & 91.91 & 90.15 & 86.76 & 81.01 & 69.77 & 54.25 & 37.55 & 17.17 & 71.44 \\
        16     & 96.22 & 95.16 & 94.43 & 92.24 & 89.17 & 79.57 & 68.14 & 47.16 & 22.77 & 4.58 & 68.94 \\
        17     & 96.36 & 95.87 & 95.26 & 93.23 & 90.65 & 87.08 & 75.10 & 59.85 & 38.74 & 11.36 & 74.35 \\
        18     & 96.82 & 96.51 & 96.23 & 95.31 & 92.84 & 84.06 & 73.08 & 54.41 & 26.99 & 3.95 & 72.02 \\
        19     & 96.51   & 95.82 & 95.44 & 94.23 & 92.00 & 86.98 & 80.26 & 62.57 & 41.40 & 29.00 & 77.42 \\
        20     & 97.48 & 97.33 & 96.77 & 92.02 & 89.02 & 84.29 & 76.69 & 60.04 & 41.98 & 18.64 & 75.43 \\
        21     &  96.67 &   96.23      &    95.68    &    94.53     &   92.02     &     88.66    &    77.92    &     63.96    &  46.53      &   35.13      &  78.74 \\ \hline
        \end{tabular}
        \caption{The XRF V2 dataset allows combinations of different devices, and we tested the most common device combinations, including individual devices and their configurations.}
        \label{fig:devices-combination}
    \end{minipage}
\end{figure}

\begin{figure}[!ht]
    \centering
    \includegraphics[width=0.95\linewidth]{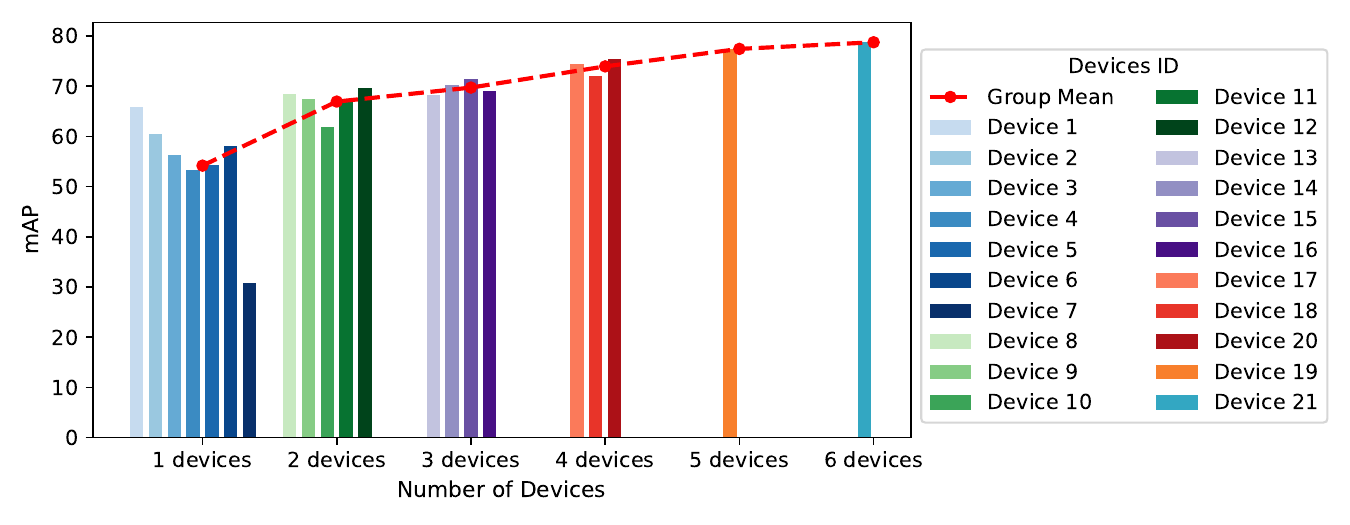}
    \caption{Device combinations results. As more devices are added, XRFMamba's performance improves.}
    \label{fig:device-combinations-number-of-devices}
\end{figure}

\subsubsection{Device Combination Study}\label{sec:devive-combination}

XRF V2 dataset includes Wi-Fi signals and IMU data from seven types of devices: phones, watches, earbuds, and glasses, as shown in Fig.\ref{fig:devices-combination}. Users can combine devices according to their own needs. In Fig.~\ref{fig:devices-combination}, we report some of the most common device combinations, such as case 9 (a phone and a watch), case 14 (a phone, a watch, and an earbud), case 17 (a phone, a watch, glasses, and ambient Wi-Fi). It shows that the average mAP of using only a right-hand-worn smartwatch (case 1) is 65.81, and case 17 achieves an average mAP of 74.35.
These results also have important implications for practical applications. Users can combine devices according to their preferences, and the system can adapt to different effective and valid device combinations for data input. For example, if a smartphone's IMU malfunctions or is not worn as intended (e.g., left on a table), the IMU data may no longer accurately represent the user's actions. Similarly, a malfunctioning Wi-Fi router can disrupt Wi-Fi signals. This flexibility ensures that the system remains functional even when some sensors are not operating optimally.

Fig.~\ref{fig:device-combinations-number-of-devices} shows the model's performance as the number of devices increases. The x-axis represents the number of devices involved in each combination, and each bar corresponds to the mAP value achieved by XRFMamba for that specific configuration. As illustrated in the figure, the performance of XRFMamba consistently improves with the inclusion of more devices. This trend suggests that the model benefits from the increased data diversity provided by multiple devices, which enhances its robustness and accuracy in recognizing user action sequences. The results highlight XRFMamba's strong scalability to generalize effectively in real-world scenarios involving heterogeneous device environments.

\begin{figure}[ht]
    \centering
    \includegraphics[width=0.7\linewidth]{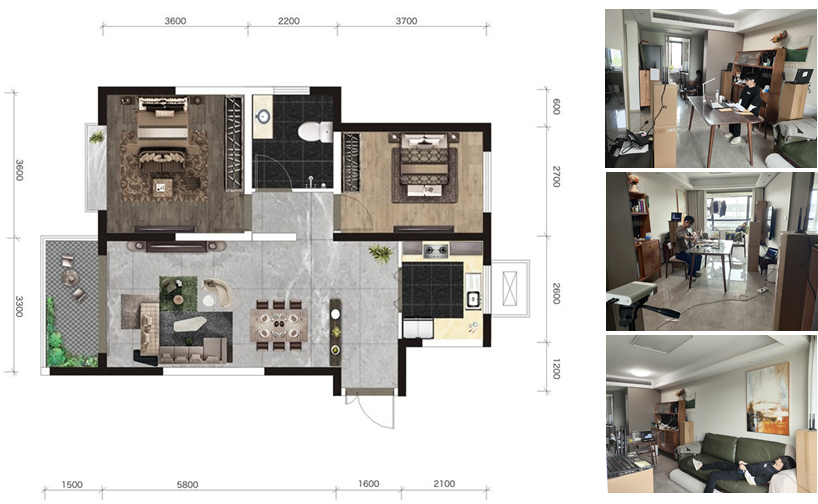}
    \caption{We recruited 5 volunteers to conduct a zero-shot evaluation in a real apartment.}
    \label{fig:real-world}
\end{figure}

\subsection{A Real-world Zero-shot Evaluation}\label{sec:real-world}

We further conduct a zero-shot evaluation in a real apartment (as shown in Fig.~\ref{fig:real-world}) to demonstrate the potential of our system for action localization and summarization in real-life settings. We recruited 5 volunteers, each in the living room, performing 15 different action sequences corresponding to the bedroom, study room, and dining room. During the actions, we collected Wi-Fi and IMU data. We then used the XRFMamba model, trained on the XRF V2 dataset, to infer directly on the new apartment data without any fine-tuning. The results of temporal action localization are presented in Table~\ref{tab:user-real-world}. As shown, the five participants achieved around 70 mAP@avg for the action sequences, showcasing the strong real-world deployment potential and value of our system.

\begin{table}[ht]
\centering
\caption{In a real apartment, XRFMamba achieves approximately 70 mAP@avg for zero-shot temporal action localization across five new volunteers, demonstrating the strong real-world deployment potential of our system. Scene abbreviations: BR = Bedroom, SR = Study Room, DR = Dining Room.}
\resizebox{\textwidth}{!}{
\begin{tabular}{lcccccccccccc}
\hline
Scene & Test user & mAP@0.5 & 0.55 & 0.6 & 0.65 & 0.7 & 0.75 & 0.8 & 0.85 & 0.9 & 0.95 & mAP@avg \\ \hline
\multirow{6}{*}{BR} 
& user1 & 96.49 & 96.49 & 96.49 & 95.21 & 95.21 & 95.21 & 88.82 & 64.89 & 37.35 & 21.08 & 78.73 \\
& user2 & 86.75 & 86.75 & 86.75 & 86.75 & 80.13 & 67.63 & 58.43 & 45.38 & 21.13 & 10.87 & 63.06 \\
& user3 & 97.44 & 97.10 & 97.10 & 95.97 & 94.94 & 90.57 & 85.57 & 70.20 & 54.53 & 22.88 & 80.63 \\
& user4 & 88.48 & 88.40 & 88.38 & 88.31 & 83.25 & 80.11 & 71.17 & 62.54 & 36.88 & 30.49 & 71.80 \\
& user5 & 64.79 & 64.79 & 64.79 & 64.79 & 61.60 & 59.93 & 59.79 & 44.56 & 32.47 & 15.54 & 53.31 \\
& Average & 86.79 & 86.71 & 86.70 & 86.21 & 83.03 & 78.69 & 72.76 & 57.51 & 36.47 & 20.17 & 69.50 \\
\hline

\multirow{6}{*}{SR} 
& user1 & 96.45 & 96.45 & 96.45 & 96.45 & 95.80 & 94.17 & 94.14 & 80.07 & 56.48 & 49.45 & 85.59 \\
& user2 & 100.00 & 100.00 & 100.00 & 100.00 & 100.00 & 100.00 & 96.30 & 76.30 & 58.52 & 35.19 & 86.63 \\
& user3 & 85.45 & 83.56 & 81.73 & 75.72 & 75.70 & 73.64 & 66.75 & 58.85 & 32.23 & 29.56 & 66.32 \\
& user4 & 94.66 & 92.73 & 90.69 & 90.41 & 88.69 & 78.76 & 74.98 & 62.84 & 42.11 & 29.48 & 74.54 \\
& user5 & 76.94 & 76.30 & 69.44 & 67.41 & 58.07 & 50.12 & 33.58 & 21.37 & 9.39 & 5.32 & 46.79 \\
& Average & 90.70 & 89.81 & 87.66 & 86.00 & 83.65 & 79.34 & 73.15 & 59.89 & 39.75 & 29.80 & 71.97 \\
\hline

\multirow{6}{*}{DR} 
& user1 & 89.79 & 89.79 & 89.19 & 89.19 & 88.00 & 87.09 & 77.24 & 60.59 & 47.61 & 34.07 & 75.26 \\
& user2 & 91.09 & 89.85 & 89.75 & 89.61 & 83.78 & 83.70 & 73.31 & 57.66 & 47.39 & 34.79 & 74.09 \\
& user3 & 92.67 & 92.67 & 91.53 & 91.53 & 87.98 & 87.83 & 84.50 & 68.55 & 62.97 & 47.27 & 80.75 \\
& user4 & 91.41 & 90.56 & 80.92 & 77.62 & 61.64 & 59.21 & 47.62 & 37.60 & 20.07 & 13.47 & 58.01 \\
& user5 & 85.33 & 85.33 & 85.33 & 85.33 & 82.60 & 73.86 & 67.80 & 60.87 & 35.17 & 23.51 & 68.51 \\
& Average & 90.06 & 89.64 & 87.34 & 86.66 & 80.80 & 78.34 & 70.09 & 57.05 & 42.64 & 30.62 & 71.32 \\
\hline
\end{tabular} }
\label{tab:user-real-world}
\end{table}

\section{Dataset and Code Availability}\label{sec:code}

The data and code are available at~\href{https://github.com/aiotgroup/XRFV2}{\textcolor[RGB]{255, 69, 69}{https://github.com/aiotgroup/XRFV2}}. 
The repository includes a comprehensive $\mathtt{readme.md}$ file, which describes how to use the code to reproduce our results. The code features a well-structured modular design, making it easy to call and test various backbones and modality combinations. In this code repository, under the \textit{/model} directory, you will find different method implementations, including XRFMamba and all comparison methods. You can easily use the model you need through \textit{make\_model()} found in \textit{/model/models.py}. The code for XRFMamba is located in \textit{model/TAD\_multi\_fusion.py}. Data loading methods are defined in the \textit{/dataset} directory. To get started, you need to modify the paths in \textit{basic\_config.json} to match your system setup. Then, run \textit{script/train\_run.py} to train the model. Finally, you can evaluate the model by running \textit{script/test\_run.py}. For more details, please check out the repository.

\section{Potential Usefulness and Limitations}\label{sec:usefulness-limitations}

\subsection{Potential Usefulness}\label{sec:usefulness}

XRF V2 is the first dataset designed to support temporal action localization (TAL) using Wi-Fi and IMU data, integrated with large language model (LLM) agents for action summarization in the smart-home context. We believe the XRF V2 dataset can significantly contribute to the community in at least the following areas.

\subsubsection{Advanced Methods}

\begin{enumerate}
    \item  The XRF V2 dataset can serve as a benchmark for action recognition, temporal action localization, and action summarization based on smartphones, watches, earbuds, glasses, and 
    Wi-Fi. It supports research aimed at developing methods that either achieve higher precision or are more lightweight.

    \item XRF V2 also supports weakly supervised action localization, where only the action categories are provided without specifying the precise start and end times of the actions. This setup allows for research on action localization methods that can infer temporal boundaries from less detailed supervision, enabling more scalable and practical approaches for real-world applications.

    \item The current action summarization is two-stage, consisting of the first stage, XRFmamba, and the second stage, the LLM agent. Future advanced methods could explore LLMs handling Wi-Fi and IMU time series, including Wi-Fi and IMU data tokenization, alignment with LLMs, and LLM fine-tuning, to enable end-to-end action summarization.
    
\end{enumerate}

\begin{figure}[ht]
    \centering
    \includegraphics[width=1\linewidth]{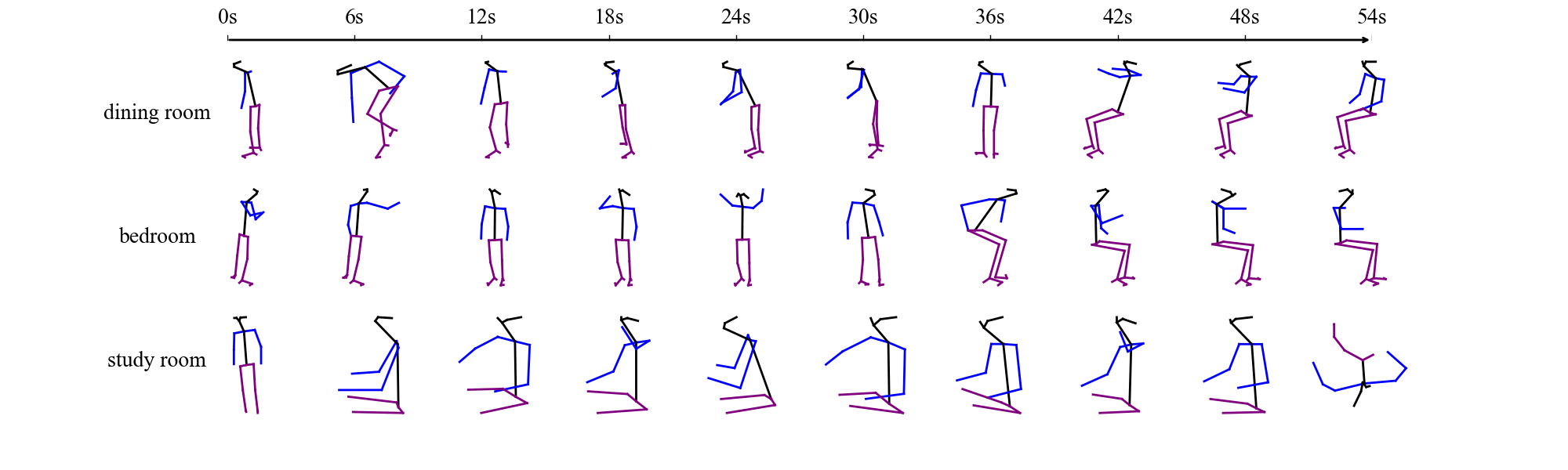}
    \caption{XRF V2 already includes 2D pose data, processed using OpenPose's Body25 model~\cite{cao2019openpose} on Kinect videos. This allows for 2D pose estimation, tracking, and prediction. }
    \label{fig:2dpose-future-work}
\end{figure}

\subsubsection{More Tasks}

\begin{enumerate}
    \item  XRF V2 supports the research of next action prediction, which facilitates the development of systems that can anticipate and respond to user actions. For example, by detecting the user’s intent to turn on the lights, the system can automatically trigger the action without the need for voice or manual control. This capability enhances the intelligence of smart devices, making them more context-aware and responsive.

     \item XRF V2 already includes 2D pose data, processed using OpenPose's Body25 model on Kinect videos~\cite{cao2019openpose}, as shown in Fig.~\ref{fig:2dpose-future-work}. This allows for 2D pose estimation, tracking, and prediction based on WiFi, IMU, and video data. Compared to previous datasets such as Person-in-WiFi 3D~\cite{yan2024person} and MM-Fi~\cite{yang2024mm}, where the person stands at a fixed position or repeats actions, the actions in XRF V2 are more representative of real-world scenarios, capturing more dynamic and varied human behavior.  
    
    \item In the future, we plan to gradually extend the dataset to include 3D human pose, 3D human mesh, and densepose data, enabling more fine-grained human sensing research.
    
\end{enumerate}

\subsubsection{More Research and Industry Exploration}

\begin{enumerate}
    \item XRF V2 supports multimodal learning between IMU, WiFi, and video, facilitating the development of multimodal foundation models. While the original videos cannot be shared due to privacy concerns with volunteers, we are to process them into features using InterVideo-6B~\cite{Internvideo} and make these features publicly available.

    \item Realistic data generation is crucial in reducing the need for data collection and domain generalization. XRF V2 is a dataset collected using real devices, and it provides high-quality data labels. It can be used to promote or evaluate research on IMU and Wi-Fi data generation methods.

    \item The academic and industrial sectors are welcome to leverage our dataset to develop specific applications, such as ambient sensing systems and context-aware systems, which help in enhancing smart home automation, health, and wellness monitoring.
    
\end{enumerate}

\subsection{Limitations and Future Work}\label{sec:limitations}

XRF V2 also presents several limitations that future work can address:

\begin{enumerate}
    \item XRF V2 currently includes only the bedroom, dining room, and study room. However, there are many other significant environments, such as the kitchen, bathroom, and living room, which are also crucial for ambient sensing. Future work will expand XRF V2 by collecting data from a wider variety of settings, thereby enriching the dataset. All new data will be made publicly available on the XRF V2 project page.

    \item  Currently, XRF V2 uses a sequential labeling strategy, where actions are performed one after another, with each time segment annotated with a single dominant action. However, the current XRF V2 does not provide concurrent action annotations, such as walking while answering the phone, which limits the dataset’s ability to fully represent complex user behaviors. To enhance the dataset, future research could focus on capturing concurrent action scenarios and proposing methods for detecting such concurrent actions.

    \item Currently, XRF V2 is designed for single-user scenarios, but multi-user scenarios are also common in real-life settings. Future work can expand the dataset and methods to accommodate multiple users. Since IMUs are worn by the users, they inherently support multi-person sensing. As for Wi-Fi, our approach can explore the integration of multi-person Wi-Fi sensing methods, such as WiMesh~\cite{wang2022wi}, Person-in-WiFi~\cite{wang2019person, yan2024person}, and Densepose from Wi-Fi~\cite{geng2022densepose}, all of which have demonstrated the ability to distinguish multiple individuals using Wi-Fi signals.

    \item XRF V2 currently uses a specific 1-transmitter-3-receiver configuration. Future work can explore configurations with fewer receivers, such as 1-receiver or 2-receiver setups, to optimize deployment for real-world scenarios. Additionally, future research can investigate the generalization capability of the system under varying device deployment scenarios. For example, we can treat a Tx-Rx\#1 setup (1-transmitter-1-receiver) as the source domain and a Tx-Rx\#2 setup (another 1-transmitter-1-receiver configuration) as the target domain, enabling the study of the system’s adaptability across different setups.
 
\end{enumerate}

\section{Conclusion}\label{sec:conclusion}

We present the XRF V2 dataset, which captures continuous action sequences from 16 volunteers across three typical household environments: the dining room, study room, and bedroom. Comprising a total of 825 action sequences, the dataset integrates data from various sensors, including IMUs from smartwatches, smartphones, earbuds, and glasses, Wi-Fi transceivers, and synchronized video from an Azure Kinect. The diverse and personalized action sequences, along with synchronized multimodal data, provide a valuable resource for advancing research in human action understanding and multimodal machine learning. We also introduce a novel human action understanding task: action summarization using Wi-Fi and IMU signals in smart-home environments, for which we design a new evaluation metric, response meaning consistency~(RMC). Action summarization takes action sequences and task-oriented prompts as input to large language models (LLMs), transforming the LLMs into intelligent agents. We demonstrate that with IMU or Wi-Fi input, LLMs can function as personal assistants, health assistants, and home assistants. Furthermore, we propose the XRFMamba neural network, which effectively performs temporal action summarization and action summarization, and conduct a detailed evaluation showing its superior performance compared to existing state-of-the-art methods.

\section*{Acknowledgements}

This work was supported by the National Natural Science Foundation of China under grants U21A20462, 62372400, 62372365, and 62472346, and
Fundamental Research Funds for the Central Universities. We are grateful to anonymous Associate Editors and Reviewers for their invaluable comments, revision suggestions, and recommendation. We thank all volunteers and human auditors for their participation.

\bibliographystyle{ACM-Reference-Format}
\bibliography{refrence}

\appendix

\section{Values of Proposed Actions}\label{sec:action-descript-and-value}

\begin{longtable}{|p{3cm}|l|p{3.5cm}|p{5.5cm}|}
\caption{Action descriptions and the use value.}
\label{tab:all-action-values}
\\
\hline
\textbf{Action Name} & \textbf{Scene} & \textbf{Description} & \textbf{Utility Value} \\
\hline
\endfirsthead
\multicolumn{4}{c}%
{{\bfseries \tablename\ \thetable{} -- continued from previous page}} \\
\hline
\textbf{Action Name} & \textbf{Scene} & \textbf{Description} & \textbf{Utility Value} \\
\hline
\endhead
\hline \multicolumn{4}{r}{{Continued on next page}} \\ \hline
\endfoot
\hline

\endlastfoot

\rowcolors{2}{blue!5}{white} 
{Walk} & {Study room} & {Moving position. }& {Health Assistant: Tracks users exercise time and conducts exercise monitoring.}\\
\hline
{Sit down} & {Study room} & {Sitting on a chair/bed/sofa. } & {Health Assistant: Monitors users' sedentary time and reminds them to engage in activities.}\\
\hline
{Stand up} & {Study room }& {Getting up from the chair/bed/sofa.} &{ Health Assistant: Monitors whether users are sitting safely (provides timely feedback in case of falls).}\\
\hline
{Pour water into the cup} & {Study room }& {Transferring water from a water source (such as a pitcher or faucet) into a cup. }& {Health Assistant: Helps monitor users' water intake and frequency .}\\
\hline
{Drink water }& {Dining room} & {Consuming water from a cup or glass.} & {Health Assistant: Tracks hydration and reminds users to drink water for health monitoring.} \\
\hline
{Take medicine }& {Dining room }& {Consuming prescribed medicine from a container. } & {Health Assistant: Reminds and tracks whether a person has taken their medication. }\\
\hline
{Pick up things } & {Dining room} & {Grasping and lifting objects from the floor. } & {Home Assistant: Monitors object interaction and movement; can help in detecting potential clutter or safety hazards. } \\
\hline
{Take the fruits from the cabinet } & {Dining room } & {Retrieving fruits stored in a cabinet.} & {Health Assistant: Tracks food access related to healthy eating habits.}\\
\hline
{Cut fruits } & {Dining room} & {Using a knife or other cutting tools to divide fruits into smaller pieces. } & {Health Assistant: Monitors food preparation related to healthy eating.}\\
\hline
{Eat fruits} & {Dining room } & {Consuming fruits as a part of a meal or snack.} & {Health assistant: Tracks the frequency and time of users' meals. }\\
\hline
{Wash hands} & {Dining room} & {Washing hands with water.} & {Health Assistant: Encourages hygiene practices, especially before eating or after handling food.} \\
\hline
{Throw waste} & {Dining room} & {Disposing of trash or unwanted items in a waste bin. } & {Home Assistant: Monitors waste management and can be used to promote proper waste disposal habits.}\\
\hline
{Wipe the table} &  {Dining room} & {Cleaning the surface of the table using a cloth or other cleaning tools. }  & {Home Assistant: Monitors cleaning activities and can be used to maintain a clean living environment.}\\
\hline
{Stretching} & {Dining room} & {Perform stretching exercises to achieve the goal of relaxing the body. }  & {Health Assistant: Tracks physical activity related to flexibility and muscle health; can also be used to remind users to stretch during sedentary periods.}\\
\hline
{Turn on and off the desk lamp} & {Study room}  & {Switching the desk lamp on or off.} & {Personal Assistant: Provides environmental control, helping with lighting adjustments during work or study. } \\
\hline
{Operate the mouse}  & {Study room} & {Manipulating a computer mouse to control the cursor and perform actions on a computer screen.} & {Personal assistant: tracks user interaction with the computer and can be used to evaluate user work hours.}\\
\hline
{Write} & {Study room} & {Using a pen, pencil, or other writing utensils to put words or marks on paper or a writing surface.} & {Personal assistant: Supervise creative or work-related writing activities; Can be used to track work hours such as note taking.}\\
\hline
{Operate the keyboard} & {Study room} & {Typing or interacting with a keyboard for work or communication.} & {Personal Assistant: Tracks user productivity and interaction with the computer. } \\
\hline
{Read a book} & {Study room}  & {Perusing the pages of a book, absorbing information or for leisure.} & {Personal assistant: Tracks reading habits and can be used to monitor reading time.}\\
\hline
{Open an envelope} & {Study room} & {Tearing or cutting open an envelope to access its contents.} & {Personal assistant: Monitor email-related activities; Can be used to remind users to view and process received letters at fixed times.}\\
\hline
{Answer the phone} & {Study room} & {Picking up the phone to receive a call.} & {Personal Assistant: Assists with communication and task management during phone calls. }\\
\hline
{Write on the blackboard} & {Study room}  & {Using chalk or a marker to write or draw on a blackboard, often for educational or presentational purposes.} & {Personal assistant: Can be used to monitor user behavior related to blackboard writing.}\\
\hline
{Get up} & {Bedroom} & {Rising from a lying or sitting position.} & {Home Assistant: Monitors wake-up and activity times, helps track sleep or mobility patterns. } \\
\hline
{Lie down} & {Bedroom} & {Lying down on a bed, typically for rest or sleep.} & {Home Assistant: Detects sleep mode initiation and monitors sleep quality and patterns. } \\
\hline
{Use phone}  & {Bedroom}  & {Interacting with a mobile phone, such as making calls, sending messages, or using apps.} & {Personal assistant: Monitors habits of using mobile phones; Can be used to monitor the time spent on mobile phones.}\\
\hline
{Open and close curtains}  & {Bedroom}  & {Pulling or pushing curtains to cover or uncover a window.} & {Home Assistant: Monitors users' habits regarding lighting requirements and assists users in managing the switch of curtains}\\
\hline
{Water plants}  & {Bedroom} & {Pouring water on plants to keep them hydrated.} & {Home Assistant: Monitors plant-care activities; can be used to ensure the well-being of indoor plants.}\\
\hline
{Stand still}  & {Bedroom} & {Remaining in a stationary standing position.} & {Health assistant: Can be used to monitor users' behavioral habits}\\
\hline
{Lying still} & {Bedroom} & {Remaining motionless while lying down, often during rest or sleep.} & {Home assistant: by detecting the user's static time lying in bed; It can also be used to monitor users' behavior habits regarding rest/sleep. And it can remind users to rest in a timely manner}\\
\hline
{Open and close windows} & {Bedroom} & {Opening or closing a window to adjust airflow or lighting.} & {Home Assistant: Assists in managing environmental factors like air quality and room temperature. } \\
\hline
\end{longtable}

\section{ Volunteer Coordination and Action Execution}\label{sec:volunteer}

We recruit volunteers through group chats, online meetings, and social media. Volunteers can access the informed consent form online, which includes details about the study's purpose, methods, participation duration, potential risks, and compensation. Registrants receive a shared editable online spreadsheet to fill in their preferred participation time. At the experiment site, volunteers are provided with a printed version of the informed consent form. We explain its content to each volunteer and address questions they might have. If volunteers agree to proceed, they sign to confirm their participation and receive compensation equivalent to the local average wage multiplied by their participation duration.

Before starting the experiment, we assist volunteers in wearing the IMU devices and explain the data collection process and relevant precautions. Once the volunteers feel ready, the experiment begins. Volunteers still retain the right to withdraw from the experiment at any time. Our experiment is conducted with approval from the Institutional Review Board (IRB). We recruit 16 volunteers, with an age range of 22–34 years, a height range of 1.57–1.82 meters, and a weight range of 43–90 kgs.

Before performing the action sequences, we proportionally shorten the durations estimated by the volunteers to fall within 5–20 seconds. This adjustment is necessary because the originally estimated durations are often much longer; for instance, reading a book might be estimated as one hour. Performing such lengthy actions would significantly increase data collection time and complicate annotation. For example, it is challenging for individuals to maintain a single activity, such as reading, for an extended period without engaging in other incidental actions, such as adjusting their posture or picking up a drink. When these additional actions occur, the annotation process becomes more complicated, as it requires meticulously reviewing the entire video segment to precisely identify and mark the start and end times of each action. 

We also make reasonable adjustments to the execution durations several times, generating more action sequences. As a result, the action sequences in each scene become more diverse. After that, we can adopt an automated annotation method to streamline the process. The action sequences proposed by the volunteers are converted into audio prompts using text-to-audio technology. These audio prompts are played in real time to guide the volunteers. Upon hearing an action instruction, the volunteer performs the corresponding action and transitions to the next action upon hearing the subsequent prompt. This process continues until the entire sequence is completed.

\section{Decomposed Bidirectionally Mamba Block Process}\label{sec:dbm} 

\begin{algorithm}[ht]
\caption{Decomposed Bidirectionally Mamba Block Process (in ViM~\cite{zhu2024vision} style)}
\small
\begin{algorithmic}[1]
\REQUIRE{sensory sampling point $\mathbf{S}_{l-1}$ : \textcolor{shapecolor}{$(\mathtt{B}, \mathtt{L}, \mathtt{D})$}}
\STATE \textcolor{gray}{\text{/* normalize the input sequence $\mathbf{S}_{l-1}'$ */}}
\STATE $\mathbf{S}_{l-1}'$ : \textcolor{shapecolor}{$(\mathtt{B}, \mathtt{L}, \mathtt{D})$} $\leftarrow$ $\mathbf{LayerNorm}(\mathbf{S}_{l-1})$
\STATE $\mathbf{z1}$ : \textcolor{shapecolor}{$(\mathtt{B}, \mathtt{L}, \mathtt{E})$} $\leftarrow$ $\mathbf{Linear}^\mathbf{1}(\mathbf{S}_{l-1}')$
\STATE $\mathbf{x_{forward}}$ : \textcolor{shapecolor}{$(\mathtt{B}, \mathtt{L}, \mathtt{E})$} $\leftarrow$ $\mathbf{Linear}^\mathbf{2}(\mathbf{S}_{l-1}')$
\STATE $\mathbf{x_{backward}}$ : \textcolor{shapecolor}{$(\mathtt{B}, \mathtt{L}, \mathtt{E})$} $\leftarrow$ $\mathbf{Linear}^\mathbf{3}(\mathbf{S}_{l-1}')$
\STATE $\mathbf{z2}$ : \textcolor{shapecolor}{$(\mathtt{B}, \mathtt{L}, \mathtt{E})$} $\leftarrow$ $\mathbf{Linear}^\mathbf{4}(\mathbf{S}_{l-1}')$
\STATE \textcolor{gray}{\text{/* process with different direction */}}
\FOR{$o$ in \{forward, backward\}}
\STATE $\mathbf{x}'_o$ : \textcolor{shapecolor}{$(\mathtt{B}, \mathtt{L}, \mathtt{E})$} $\leftarrow$ $\mathbf{SiLU}(\mathbf{Conv1d}_o(\mathbf{x_o}))$
\STATE $\mathbf{B}_o$ : \textcolor{shapecolor}{$(\mathtt{B}, \mathtt{L}, \mathtt{N})$} $\leftarrow$ $\mathbf{Linear}_o^{\mathbf{B}}(\mathbf{x}'_o)$
\STATE $\mathbf{C}_o$ : \textcolor{shapecolor}{$(\mathtt{B}, \mathtt{L}, \mathtt{N})$} $\leftarrow$ $\mathbf{Linear}^{\mathbf{C}}_o(\mathbf{x}'_o)$
\STATE \textcolor{gray}{\text{/* softplus ensures positive $\Delta_o$ */}}
\STATE $\Delta_o$ : \textcolor{shapecolor}{$(\mathtt{B}, \mathtt{L}, \mathtt{E})$} $\leftarrow$ $\log(1 + \exp(\mathbf{Linear}_{o}^{\Delta}(\mathbf{x}'_o) + \mathbf{Parameter}_o^{\Delta}))$
\STATE \textcolor{gray}{\text{/* shape of $\mathbf{Parameter}_o^{\mathbf{A}}$ is \textcolor{shapecolor}{$(\mathtt{E}, \mathtt{N})$} */}}
\STATE $\overline{\mathbf{A}_o}$ : \textcolor{shapecolor}{$(\mathtt{B}, \mathtt{L}, \mathtt{E}, \mathtt{N})$} $\leftarrow$ $\Delta_o \bigotimes \mathbf{Parameter}_o^{\mathbf{A}}$ 
\STATE $\overline{\mathbf{B}_o}$ : \textcolor{shapecolor}{$(\mathtt{B}, \mathtt{L}, \mathtt{E}, \mathtt{N})$} $\leftarrow$ $\Delta_o \bigotimes \mathbf{B}_o$
\STATE \textcolor{gray}{\text{/* initialize $h_o$ and $\mathbf{y}_o$ with $0$ */}}
\STATE $h_o$ : \textcolor{shapecolor}{$(\mathtt{B}, \mathtt{E}, \mathtt{N})$} $\leftarrow$ zeros \textcolor{shapecolor}{$(\mathtt{B}, \mathtt{E}, \mathtt{N})$}
\STATE $\mathbf{y}_o$ : \textcolor{shapecolor}{$(\mathtt{B}, \mathtt{L}, \mathtt{E})$} $\leftarrow$ zeros \textcolor{shapecolor}{$(\mathtt{B}, \mathtt{L}, \mathtt{E})$}
\STATE \textcolor{gray}{\text{/* SSM recurrent */}}
\FOR{$i$ in \{0, ..., M-1\}}
\STATE $h_o$ = $\overline{\mathbf{A}_o}[:,i,:,:] \bigodot h_o + \overline{\mathbf{B}_o}[:,i,:,:] \bigodot \mathbf{x}_o'[:,i,:,\textcolor{shapecolor}{\mathtt{None}}] $
\STATE $\mathbf{y}_o[:,i,:]$ = $h_o \bigotimes \mathbf{C}_o[:,i,:]$
\ENDFOR
\ENDFOR
\STATE \textcolor{gray}{\text{/* get gated $\mathbf{y}$ */}}
\STATE $\mathbf{y_{forward}^{'}}$ : \textcolor{shapecolor}{$(\mathtt{B}, \mathtt{L}, \mathtt{E})$} $\leftarrow$ $\mathbf{y_{forward}} \bigotimes \mathbf{SiLU}(\mathbf{z1}) $
\STATE $\mathbf{y_{backward}^{'}}$ : \textcolor{shapecolor}{$(\mathtt{B}, \mathtt{L}, \mathtt{E})$} $\leftarrow$ $\mathbf{y_{backward}} \bigotimes \mathbf{SiLU}(\mathbf{z2}) $
\STATE \textcolor{gray}{\text{/* residual connection */}}
\STATE $\mathbf{S}_{l}$ : \textcolor{shapecolor}{$(\mathtt{B}, \mathtt{L}, \mathtt{D})$} $\leftarrow$ $\mathbf{Linear}^\mathbf{S}( \mathbf{Cat} (\mathbf{y_{forward}^{'}}, \mathbf{y_{backward}^{'}})) + \mathbf{S}_{l-1}$
\STATE Return: $\mathbf{S}_{l}$ : \textcolor{shapecolor}{$(\mathtt{B}, \mathtt{L}, \mathtt{D})$}
\label{alg:dbm}
\end{algorithmic}
\end{algorithm}

\section{Statistical Significance Analysis}\label{app: Significance Analysis}

\begin{table}[ht]
\caption{Comparison of mAP performance for WiFiTAD and XRFMamba models}
\small
\begin{tabular}{ccccccccccccc}
\hline
Model & No. & mAP@0.5 & 0.55 & 0.6 & 0.65 & 0.7 & 0.75 & 0.8 & 0.85 & 0.9 & 0.95 & mAP@avg \\ \hline

\multirow{3}{*}{WiFiTAD~\cite{liu2024wificsibasedtemporal}} 
& 1 & 96.04 & 95.70 & 94.73 & 93.05 & 89.44 & 80.35 & 68.88 & 51.97 & 32.94 & 12.61 & 71.57 \\
& 2 & 94.98 & 94.52 & 90.87 & 88.61 & 85.69 & 80.87 & 74.69 & 55.77 & 35.27 & 18.33 & 71.96 \\
& 3 & 95.34    &   94.71    &   94.32     &   92.83     &   86.32     &   81.75    &    73.62    &    57.73    &  36.69      &   19.14     &  73.25 \\
& 4 & 95.22 & 94.80 & 94.12 & 92.53 & 88.76 & 81.42 & 73.29 & 55.64 & 34.82 & 15.23 & 72.58 \\
& 5 & 96.14 & 95.48 & 94.67 & 92.71 & 89.15 & 81.23 & 72.47 & 54.93 & 33.64 & 14.79 & 72.52 \\
& 6 & 97.14 & 96.85 & 96.23 & 94.91 & 91.38 & 84.54 & 75.62 & 56.48 & 37.29 & 18.16 & 74.86 \\
& 7 & 95.78 & 95.12 & 94.33 & 92.45 & 88.92 & 80.56 & 71.82 & 53.91 & 32.77 & 14.45 & 72.01 \\
& 8 & 96.78 & 96.21 & 95.62 & 94.32 & 90.89 & 83.91 & 74.67 & 57.12 & 38.46 & 19.32 & 74.73 \\
& 9 & 94.87 & 94.05 & 93.28 & 91.73 & 88.99 & 81.64 & 72.83 & 54.72 & 34.19 & 15.67 & 72.20 \\
& 10 & 96.03 & 95.68 & 95.11 & 93.87 & 90.23 & 82.96 & 74.32 & 56.11 & 36.85 & 17.42 & 73.86 \\

\hline
\multirow{3}{*}{XRFMamba} 
& 1 & 97.62 & 96.90 & 95.90 & 95.40 & 93.18 & 89.78 & 82.02 & 65.87 & 47.06 & 36.13 & 79.99 \\
& 2 & 97.53 & 97.21 & 96.89 & 96.47 & 93.54 & 90.82 & 80.36 & 64.11 & 48.30 & 38.41 & 80.36 \\
& 3 & 96.83 & 96.44 & 96.19 & 95.54 & 93.27 & 88.88 & 82.20 & 68.53 & 51.55 & 40.23 & 80.97 \\
& 4 & 97.50 & 97.15 & 96.79 & 95.83 & 92.69 & 89.70 & 78.40 & 63.70 & 46.13 & 36.22 & 79.41 \\
& 5 & 96.53 & 96.04 & 95.83 & 95.05 & 92.13 & 87.05 & 80.67 & 66.51 & 51.27 & 38.30 & 79.94 \\
& 6 & 95.56 & 94.59 & 93.82 & 93.61 & 91.08 & 86.64 & 80.77 & 62.79 & 48.55 & 38.50 & 78.59 \\
& 7 & 96.97 & 96.54 & 96.03 & 93.04 & 89.75 & 84.69 & 77.59 & 64.75 & 49.79 & 38.31 & 78.75 \\
& 8 & 96.49 & 96.23 & 93.01 & 92.00 & 89.55 & 84.88 & 78.25 & 64.22 & 46.95 & 35.18 & 77.68 \\
& 9 & 97.26 & 96.87 & 96.70 & 95.73 & 91.95 & 87.18 & 81.14 & 67.23 & 47.90 & 37.00 & 79.89 \\
& 10 & 97.04 & 96.55 & 96.35 & 95.61 & 92.82 & 88.86 & 79.81 & 65.78 & 50.71 & 38.24 & 80.18 \\

\hline
\end{tabular}
\label{tab:compare-significance}
\end{table}

\begin{table}[ht]
\centering
\caption{t-test Results Comparing XRFMamba and WiFiTAD at Different Thresholds}
\begin{tabular}{ccc}
\hline
Threshold & p-value & Conclusion \\
\hline
\text{mAP@0.5} & $2.52 \times 10^{-2}$ & \text{Significant} \\
\text{mAP@0.55} & $4.00 \times 10^{-2}$ & \text{Significant} \\
\text{mAP@0.6} & $1.10 \times 10^{-1}$ & \text{Not Significant} \\
\text{mAP@0.65} & $4.16 \times 10^{-2}$ & \text{Significant} \\
\text{mAP@0.7} & $9.17 \times 10^{-3}$ & \text{Significant} \\
\text{mAP@0.75} & $1.31 \times 10^{-4}$ & \text{Significant} \\
\text{mAP@0.8} & $2.29 \times 10^{-5}$ & \text{Significant} \\
\text{mAP@0.85} & $4.42 \times 10^{-7}$ & \text{Significant} \\
\text{mAP@0.9} & $7.66 \times 10^{-8}$ & \text{Significant} \\
\text{mAP@0.95} & $3.57 \times 10^{-10}$ & \text{Significant} \\
\text{mAP@avg} & $1.37 \times 10^{-6}$ & \text{Significant} \\
\hline
\end{tabular}
\label{tab:t-test-results}
\end{table}

We conduct a detailed statistical significance analysis comparing our method (XRFMamba) with the state-of-the-art method (WiFiTAD). Each model is trained 10 times with random initialization (without using fixed seeds) to ensure the stability and reliability of the results, as summarized in Table~\ref{tab:compare-significance}. To determine whether the performance improvements of XRFMamba over WiFiTAD are statistically significant, we perform a paired sample \textit{t}-test at each mAP@tIOU threshold (mAP@0.5, mAP@0.55, ..., mAP@0.95, and mAP@avg). We compute the \textit{t}-test using the following equation:
\begin{equation}\label{eq:t-test}
t = \frac{\bar{d}}{s_d / \sqrt{n}},
\end{equation}
where $d_i = x_i - y_i$ denotes the difference between paired results from XRFMamba ($x_i$) and WiFiTAD ($y_i$), $\bar{d} = \frac{1}{n} \sum_{i=1}^{n} d_i$ represents the mean of the paired differences, $s_d = \sqrt{\frac{1}{n - 1} \sum_{i=1}^{n} (d_i - \bar{d})^2}$ is the standard deviation of these differences, and $n$ is the number of paired results (i.e., 10 repeated trials). We calculate the \textit{p}-value based on the $t$-distribution with $n-1$ degrees of freedom, which quantifies the probability of observing such a difference (or a more extreme one) under the null hypothesis that the two methods perform equally. Generally, it is considered a statistically significant difference when $p < 0.05$.

The results are presented in Table~\ref{tab:t-test-results}. For mAP@0.6, no significant difference is found between XRFMamba and WiFiTAD, with a p-value of 0.1103. This indicates that at this specific tIoU threshold, the performance of both methods is very similar, and there is insufficient evidence to claim a statistically significant difference. However, at most other thresholds (mAP@0.5, 0.55, 0.65, 0.7, 0.75, 0.8, 0.85, 0.9, and 0.95), the performance difference is statistically significant, with p-values consistently below 0.05. Notably, at higher tIoU thresholds, such as mAP@0.75, 0.8, 0.85, 0.9, and 0.95, the p-values are extremely low, demonstrating a strong and significant performance advantage of XRFMamba over WiFiTAD. Additionally, the mAP@avg result also shows a significant improvement, with a p-value of $1.37 \times 10^{-6}$, further supporting the superior overall performance of XRFMamba. We hope these findings address your concerns and further strengthen the credibility of our results.

\section{Evaluated Task-orientated Prompts}\label{appsec:prompts} 

\subsection{An example of responses from three LLMs to the same prompt. }

\textbf{—— Input:} ``\textbf{\#\#\#\#} You are an intelligent action summarization agent. \textbf{\#\#\#\#} This is an action sequence that occurred in the bedroom: the user did the action of walking, with a start time of 0.0 and an end time of 5.0; the user did the action of sitting down, with a start time of 5.0 and an end time of 13.0; the user did the action of reading a book, with a start time of 13.0 and an end time of 22.0; the user did the action of pouring water into the cup, with a start time of 22.0 and an end time of 27.0;  the user did the action of taking medicine, with a start time of 27.0 and an end time of 40.0; the user did the action of lying, with a start time of 40.0 and an end time of 46.0; the user did the action of using the phone, with a start time of 46.0 and an end time of 60.0; the user did the action of getting up, with a start time of 60.0 and an end time of 68.0.  \textbf{\#\#\#\#} Please answer whether the user read a book before lying down during this period?''

\textbf{——Output of ChatGPT-4o:} ``Yes, the user read a book before lying down during this period. The action of reading a book occurred from 13 to 22. The action of lying down occurred from 40 to 46. Therefore, reading the book happened before lying down.''

\textbf{——Output of Qwen2.5-Plus:} ``Yes, the user read a book before lying down during this period. According to the sequence of actions: The user started reading a book at 13 and ended at 22. The user then performed other actions (pouring water, taking medicine) before starting to lie down at 40. Since the action of reading a book occurred before the action of lying down, the answer is yes, the user did read a book before lying down."

\textbf{——Output of DeepSeek-V3:} ``Yes, the user read a book before lying down during this period. The action of reading a book occurred from 13 to 22, and the action of lying down occurred from 40 to 46. Therefore, the user read a book before lying down.''

\subsection{All task-oriented questions used in action summarization evaluation.}

(1) Please answer the number of times the user has consumed water during this period?

(2) Please answer whether the user has been drinking water during this period?

(3) Please answer whether the user has taken medication during this period?

(4) Please answer the number of times the user has taken medication during this period?

(5) Please answer whether the user has been reading during this period?

(6) Please answer whether the user read a book before lying down during this period?

(7) Please answer the number of times the user watered the plants during this period?

(8) Please answer whether the user watered the plants during this period?

(9) Please answer whether the user has opened windows for ventilation during this period?

(10) Please answer whether the user was playing with their phone while lying in bed during this period?

(11) Please answer whether the user has been walking during this period?

(12) Please answer the number of times the user has stretched during this period?

(13) Please answer whether the user has eaten fruits during this period?

(14) Please answer the number of times the user has eaten fruits during this period?

(15) Please answer whether the user has wiped the table during this period?

(16) Please answer whether the user has thrown away garbage during this period?

(17) Please answer whether the user washed their hands before eating fruits during this period?

(18) Please answer whether the user washed their hands after eating fruits during this period?

(19) Please answer whether the user has washed their hands after littering during this period?

(20) Please answer whether the user has washed their hands after wiping the table during this period?

(21) Please answer how many times the user washed their hands during this period?

(22) Please answer whether the user operated the mouse during this period?

(23) Please answer whether the user operated the keyboard during this period?

(24) Please answer whether the user has opened the envelope during this period?

(25) Please answer whether the user has turned on the desk lamp during this period?

(26) Please answer whether the user answered the phone during this period?

(27) Please answer the number of times the user answered the phone during this period?

\section{Detailed Description of Three Fusion Methods}\label{appsec:fusion-codes}

In this section, we describe three different modality fusion strategies: \textbf{Gated Fusion}, \textbf{Linear Fusion}, and \textbf{Weighted Fusion}. Each method combines IMU and Wi-Fi features using different techniques. Below, we provide both the mathematical formulation and pseudocode for each method.

\subsection{Weighted Fusion}

In Weighted Fusion, each modality is given a fixed weight, and the features are combined by weighted addition. The IMU features are weighted by \( w_{\text{IMU}} = 0.8 \) and the Wi-Fi features by \( w_{\text{WiFi}} = 0.2 \) before being summed.

\textbf{Mathematical Formulation:}
\[
\mathbf{X}_{\text{IMU}}^{\text{weighted}} = w_{\text{IMU}} \times \mathbf{X}_{\text{IMU}}
\]
\[
\mathbf{X}_{\text{WiFi}}^{\text{weighted}} = w_{\text{WiFi}} \times \mathbf{X}_{\text{WiFi}}
\]
\[
\mathbf{X}_{\text{fused}} = \mathbf{X}_{\text{IMU}}^{\text{weighted}} + \mathbf{X}_{\text{WiFi}}^{\text{weighted}}
\]

\begin{algorithm}[ht]
	\caption{Weighted Fusion Implementation}
	\label{alg:weighted_fusion}
	\definecolor{codeblue}{rgb}{0.25,0.5,0.5}
	\lstset{
		backgroundcolor=\color{white},
		basicstyle=\fontsize{7.2pt}{7.2pt}\ttfamily\selectfont,
		columns=fullflexible,
		breaklines=true,
		captionpos=b,
		commentstyle=\fontsize{7.2pt}{7.2pt}\color{codeblue},
		keywordstyle=\fontsize{7.2pt}{7.2pt},
	}
\begin{lstlisting}[language=python]
def weighted_fusion(imu_features, wifi_features, w_imu=0.8, w_wifi=0.2):
    # Apply fixed weights to IMU and WiFi features
    gated_imu = w_imu * imu_features
    gated_wifi = w_wifi * wifi_features
    
    # Combine the weighted features by addition
    fused_features = gated_imu + gated_wifi
    return fused_features
\end{lstlisting}
\end{algorithm}

\subsection{Linear Fusion}

Linear Fusion applies a learned linear transformation to combine IMU and Wi-Fi features, followed by a gating mechanism. The resulting features are combined using the learned gate and passed through a fully connected layer.

\textbf{Mathematical Formulation:}
\[
\mathbf{G} = \sigma\left( \mathbf{W}_g \cdot \left( \mathbf{X}_{\text{IMU}} + \mathbf{X}_{\text{WiFi}} \right) \right)
\]
\[
\mathbf{X}_{\text{fused}} = \text{fc}\left( \mathbf{G} \times \left( \mathbf{X}_{\text{IMU}} \oplus \mathbf{X}_{\text{WiFi}} \right) \right)
\]
where \( \oplus \) denotes concatenation.

\begin{algorithm}[ht]
	\caption{Linear Fusion Implementation}
	\label{alg:linear_fusion}
	\definecolor{codeblue}{rgb}{0.25,0.5,0.5}
	\lstset{
		backgroundcolor=\color{white},
		basicstyle=\fontsize{7.2pt}{7.2pt}\ttfamily\selectfont,
		columns=fullflexible,
		breaklines=true,
		captionpos=b,
		commentstyle=\fontsize{7.2pt}{7.2pt}\color{codeblue},
		keywordstyle=\fontsize{7.2pt}{7.2pt},
	}
\begin{lstlisting}[language=python]
def linear_fusion(imu_features, wifi_features):
    # Combine IMU and WiFi features element-wise
    combined = imu_features + wifi_features

    # Apply a linear transformation to the combined features
    gate = sigmoid(linear_transformation(combined))
    
    # Concatenate the IMU and WiFi features
    concatenated_features = torch.cat([imu_features, wifi_features], dim=-1)
    
    # Apply gating mechanism (element-wise multiplication)
    gated_features = gate * concatenated_features
    
    # Pass the gated features through a fully connected layer
    fused_features = fully_connected(gated_features)
    
    return fused_features
\end{lstlisting}
\end{algorithm}

\subsection{Gated Fusion}

Gated Fusion uses a learned gating mechanism to weight the contribution of each modality. A 1D convolution is applied to the IMU and Wi-Fi features, followed by a sigmoid activation function to compute gating weights. These gated features are then combined element-wise.

\[
\mathbf{G}_{\text{IMU}} = \sigma\left( \mathbf{W}_g \cdot \mathbf{X}_{\text{IMU}} \right), \quad \mathbf{G}_{\text{WiFi}} = \sigma\left( \mathbf{W}_g \cdot \mathbf{X}_{\text{WiFi}} \right)
\]
\[
\mathbf{X}_{\text{IMU}}^{\text{gated}} = \mathbf{G}_{\text{IMU}} \times \mathbf{X}_{\text{IMU}}, \quad \mathbf{X}_{\text{WiFi}}^{\text{gated}} = \mathbf{G}_{\text{WiFi}} \times \mathbf{X}_{\text{WiFi}}
\]
\[
\mathbf{X}_{\text{fused}} = \mathbf{X}_{\text{IMU}}^{\text{gated}} + \mathbf{X}_{\text{WiFi}}^{\text{gated}}
\]

\begin{algorithm}[ht]
	\caption{Gated Fusion Implementation}
	\label{alg:gated_fusion}
	\definecolor{codeblue}{rgb}{0.25,0.5,0.5}
	\lstset{
		backgroundcolor=\color{white},
		basicstyle=\fontsize{7.2pt}{7.2pt}\ttfamily\selectfont,
		columns=fullflexible,
		breaklines=true,
		captionpos=b,
		commentstyle=\fontsize{7.2pt}{7.2pt}\color{codeblue},
		keywordstyle=\fontsize{7.2pt}{7.2pt},
	}
\begin{lstlisting}[language=python]
def gated_fusion(imu_features, wifi_features):
    # Apply 1D convolution to IMU and WiFi features
    gate_imu = sigmoid(convolution(imu_features))  
    gate_wifi = sigmoid(convolution(wifi_features))  
    
    # Apply gating mechanism (element-wise multiplication)
    gated_imu = gate_imu * imu_features  
    gated_wifi = gate_wifi * wifi_features  
    
    # Combine the gated features by addition
    fused_features = gated_imu + gated_wifi  
    return fused_features
\end{lstlisting}
\end{algorithm}


\end{document}